# Large Language Models for Extrapolative Modeling of Manufacturing Processes

Kiarash Naghavi Khanghah[1], Anandkumar Patel[2], Rajiv Malhotra[2*], Hongyi Xu[1*]

[1] School of Mechanical, Aerospace and Manufacturing Engineering, University of Connecticut, Storrs, CT 06269

[2] Department of Mechanical & Aerospace Engineering, Rutgers, the State University of New Jersey, Piscataway, NJ 08854

* Correspondence: hongyi.3.xu@uconn.edu, rm1306@soe.rutgers.edu

## Abstract

Conventional predictive modeling of parametric relationships in manufacturing processes is limited by the subjectivity of human expertise and intuition on the one hand and by the cost and time of experimental data generation on the other hand. This work addresses this issue by establishing a new Large Language Model (LLM) framework. The novelty lies in combining automatic extraction of process-relevant knowledge embedded in the literature with iterative model refinement based on a small amount of experimental data. This approach is evaluated on three distinct manufacturing processes that are based on machining, deformation, and additive principles. The results show that for the same small experimental data budget the models derived by our framework have unexpectedly high extrapolative performance, often surpassing the capabilities of conventional Machine Learning. Further, our approach eliminates manual generation of initial models or expertise-dependent interpretation of the literature. The results also reveal the importance of the nature of the knowledge extracted from the literature and the significance of both the knowledge extraction and model refinement components.

## 1. Introduction

Modeling of parametric relationships, i.e., the linkage between process parameters and performance metrics of the process or the product, is critical for control of manufacturing processes. The creation of physics-based simulations for deriving such relationships is often difficult and time-intensive (especially for novel processes) due to incomplete understanding of the underlying multiphysical multidomain interactions and associated constitutive laws. For example, existing models cannot predict the part-scale distribution of grain and void characteristics in electrically-assisted incremental forming of sheet metal due to a lack of models that can capture the non-equilibrium multiaxial thermomechanical conditions imposed on the metal (Bao et al., 2015; Chang et al., 2021; Huang & Logé, 2016; Shrivastava & Tandon, 2019). Similarly, the ICME Virtual Aluminum Castings (Allison et al., 2006) required models created iteratively across decades before being fully operational. It can also be difficult to construct physics-based models at the appropriate length and time scales. For example, such models can predict post-print sintering in Binder-jet printing on the few-particles scale (Fuchs et al., 2022; Grant et al., 2023; Mao et al., 2023; Mostafaei et al., 2021; Paudel et al., 2021; Sadeghi Borujeni et al., 2022) but cannot capture the part-scale spatial distribution of grain size, voids, cracking, and surface finish for complex geometries. Similarly, linking the nanoscale optical intensification between nanoparticle pairs to part-scale thermal modeling for millions of nanoparticles in printed circuits requires a scaling law, the formulation of which has taken many years and is still incomplete (Cleeman et al., 2022; Devaraj et al., 2020; Devaraj & Malhotra, 2019; Devaraj et al., 2021; Jahangir et al., 2020). The creation of such physics-based models is also affected by subjective interpretations of literature by the individual developing the model, so that the person-to-person variation in expertise becomes a significant factor.

Machine Learning (ML) provides an alternative by capturing complex input-output relationships without direct knowledge of the underlying mechanistic behavior. But the need for substantial experimental data constitutes a major limitation. The integration of ML with physics-based models addresses this issue. In one class of approaches low-fidelity physics-based models are used as source models and transfer learning or bias correction is applied to improve model accuracy using a small experimental dataset (Chen et al., 2023; Cleeman et al., 2023; Fernández-Godino, 2016; Kennedy & O'Hagan, 2000; Oddiraju et al., 2025). But this approach still requires a qualitatively accurate physics-based model as the source, which is again limited by human understanding and expertise. Physics-informed Neural Networks, even when complemented with experimental datasets on the domain boundaries, also have this limitation. The reason is that the prediction quality depends strongly on the form of the governing equation in the loss functions and thus on knowledge of the physics, multi-physical couplings, and underlying material behavior.

Large language models (LLMs), which have seen emergent use in diverse engineering and scientific fields (Buehler, 2024; Eslaminia et al., 2024; Naghavi Khanghah et al., 2025; Pal et al., 2024; Tian et al., 2024; Xu et al.), constitute a promising alternative. LLMs can generate analytical mathematical models (Du et al., 2024; Gong et al., 2024; Grayeli et al., 2024; Merler et al., 2024; Sharlin & Josephson, 2024; Shojaee et al., 2024) and can be considered as an effective alternative than traditional symbolic regression, which is an NP-hard problem (Virgolin & Pissis, 2022). LLMs can also learn and apply rules and exhibit reasoning power through structured prompting techniques (Wei et al., 2022; Zhu et al., 2023). LLMs augmented using Retrieval-Augmented Generation (RAG) achieve high application-specific accuracy by extracting answers from external documents based on queries. For example, RAG was used in additive manufacturing to collect operational procedures and material characteristics (Chandrasekhar et al., 2024); to uncover connections between parameters of the manufacturing process, material performance, and part surface quality (Jadhav et al., 2024). But using only RAG to generate quantitatively predictive models is subject to significant errors due to inaccuracy in retrieved response as seen in past work (Mansurova et al., 2024) and shown by the results in this paper.

In another LLM-based technique experimental datasets are used for iterative generation of analytical models, optimization of model coefficients, and assessment of the model's fit to the data (Du et al., 2024; Grayeli et al., 2024; Merler et al., 2024; Sharlin & Josephson, 2024; Shojaee et al., 2024). For example, Li et al. (Li et al., 2024) leveraged Multimodal LLMs to predict analytical equations from experimental data under user-specified constraints. But these methods depend on manual descriptions of the scientific problem and human interpretation of the literature for specification of high-performing analytical models as initial prompts. This inability to automatically leverage information embedded in the literature incurs the knowledge burden that plagues intuitive physics-based modeling, and as such this approach has not yet been used in the context of manufacturing.

This work establishes a novel LLM-based framework for automated generation of analytical models of parametric relationships in manufacturing processes. The novelty lies in integrating RAG-based knowledge retrieval for human-interpretation-free extraction of information from the literature with automated LLM-driven model refinement for enhancing model accuracy beyond the capabilities of RAG-only methods.

Three processes with different principles of operations are used as testbeds to examine the efficacy of our approach. The results are also used to reveal whether knowledge retrieval should extract equations or descriptions of parametric relationships from the literature, to study the significance of the model refinement and knowledge retrieval components, and to compare the performance of our framework to traditional ML with the same small experimental dataset. The predictive accuracy of the models is examined in the context of extrapolation beyond the range of input variables used for the training dataset. Section 2 introduces the proposed framework. Section 3 describes the testbed processes and section 4 presents the results. Section 5 summarizes the key observations and their implications.

## 2. Methodology

### 2.1 Overview of the framework

The traditional approach to generating physics-based models relies on human interpretation of problems using insights from related literature and prior knowledge (Figure 1a). This method often involves a trial-and-error process to achieve acceptable results, making it prone to human error and particularly challenging for high-dimensional problems. With advancements in machine learning, data-driven models have emerged as an alternative (Figure 1b). However, these models typically function as black boxes, lacking interpretability and struggling with extrapolation (Muckley et al., 2023). Additionally, they do not effectively incorporate well-established physics-based knowledge, relying solely on the provided data.

To address these limitations, we propose an LLM-based framework that integrates the strengths of both approaches (Figure 1c). Our framework consists of two key components designed to enhance model interpretability while maintaining the adaptability of data-driven methods. The first component uses RAG to retrieve knowledge on parametric relationships, including textual descriptions or equations, from literature that is potentially related to the target manufacturing process. A detailed explanation of the first component is provided in Section 2.2. The second component involves an LLM that utilizes this retrieved information to generate initial mathematical models. Furthermore, An iterative refinement process, inspired by the work of Shojaee et al. (Shojaee et al., 2024), uses LLMs to refines the equations and enhance their accuracy based on a small experimental dataset. This is possible due to the ability of LLMs to utilize previously generated answers as a hint to improve accuracy at each iteration (Zheng et al., 2023). A detailed explanation of the second component is provided in Section 2.3. The knowledge retrieval component eliminates the need for human interpretation of the literature while the refinement component addresses the known accuracy limitations of LLM methods that are based purely on knowledge retrieval. Existing approaches that generate models without incorporating the knowledge retrieved in the first component (Without RAG) will be compared with our proposed framework (With RAG) in the results section.

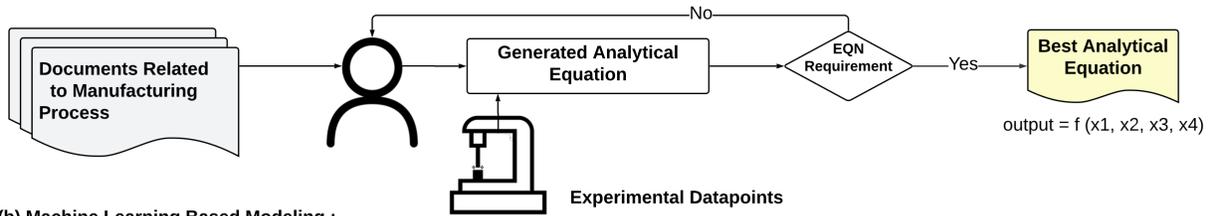

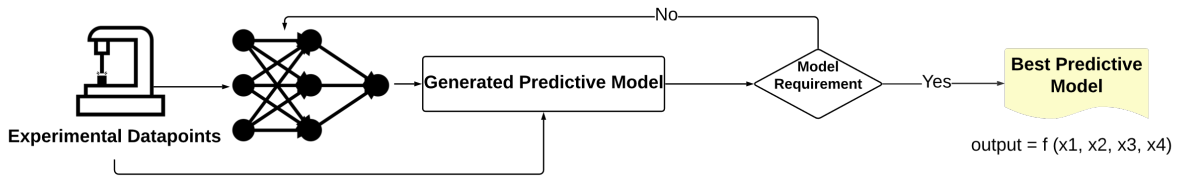

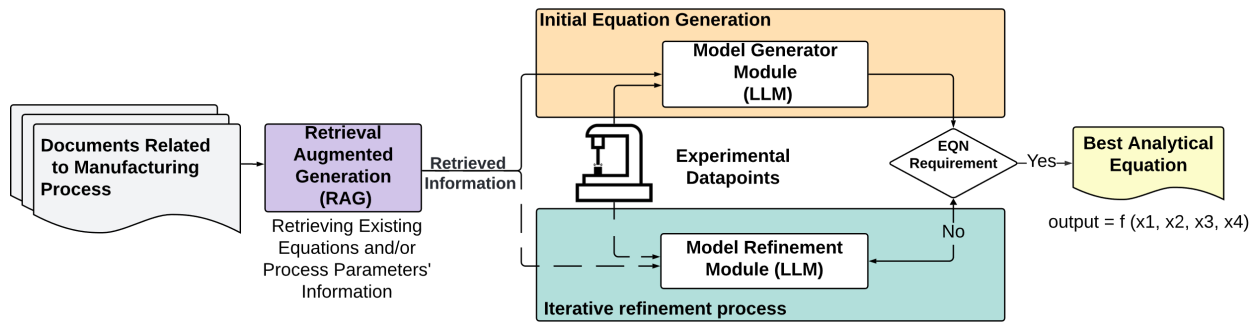

Figure 1: The flowchart of the proposed framework.

## 2.2. Knowledge retrieval

RAG is employed to process and retrieve information on parametric relationships from the literature. As shown in Figure 2, research papers (in PDF format) for processes that are potentially related to the process of interest are collected and processed using Llamaparse (*LlamaParse*). This parses these paper's content into manageable chunks (Sharma et al., 2024), i.e., smaller and more manageable sections, by dividing each larger document into parts. Each chunk is embedded via OpenAI's "text-embedding-3-small" model , thus transforming the text into vector representations for similarity-based retrieval. For the process of interest, a query is designed to analyze these vector representations to discern the relationships between process parameters and the resultant process or product performance characteristics of interests, including the qualitative nature of these relationships (e.g., a quadratic increase in output "A" due to input "B").

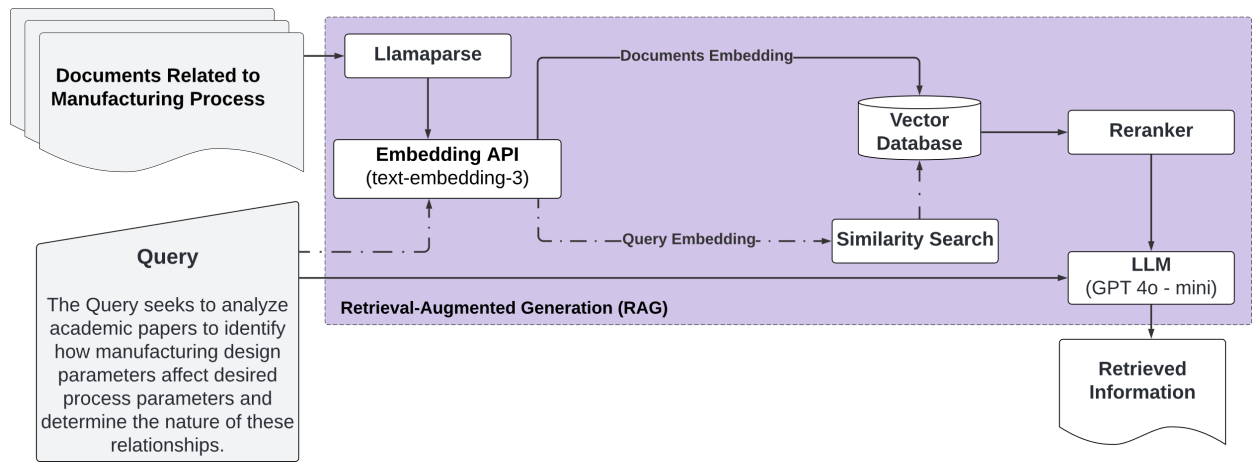

Figure 2: Flowchart for knowledge retrieval component.

Two query forms, shown in Table 1 and inputted into RAG through the Query section as illustrated in Figure 2, are created for knowledge extraction. The first form (Query form 1) retrieves textual descriptions of parametric relationships and summarizes them. The second query form (Query form 2) aims to identify and extract any existing equations. This dual-query approach ensures that both descriptive and mathematical representations of the parametric relationships are extracted from the literature. This query approach is used to perform a semantic search against the knowledge base to retrieve the most relevant chunks obtained from parsing. Re-ranking of the top-*k* chunks ensures that knowledge retrieval utilizes information that is most relevant to parametric relationships for the process at hand (Ampazis, 2024; Glass et al., 2022; Moreira et al., 2024). These chunks are then used by the LLM (OpenAI's GPT-4o-mini (OpenAI)) to define the relationships between process parameters and process performance metrics.

Table 1: Query forms for knowledge retrieval.

| Query Form 1 Analysis of Variable Relationships in PDF Documents | Query Form 2 Equation Finding and Analysis of Variable Relationships in PDF Documents |
|---|---|
| **Input:** PDF documents containing academic papers, Input variables, output process variable<br>**Output:** Relationships between input variables and output process variable<br>Use the information from PDF documents<br>**for** all input\_variables **do**<br>    Analyze relationship with output process variable<br>    **if** relationship is identified **then**<br>        Store relationship type (e.g., positive, negative, logarithmic, polynomial, exponential, etc.)<br>    **end if**<br>    Retrieve the influence of each input variable on the output process variable<br>    Retrieve relationship type (e.g., linear, nonlinear, logarithmic, etc.)<br>    **for** all observed relationship **do**<br>        Retrieve relevant tables or text descriptions as evidence<br>    **end for**<br>    Include explanations of process and variables<br>**end for** | **Input:** PDF documents containing academic papers, Input variables, output process variable<br>**Output:** Equations for output process variable as a function of input variables; relationships between input variables and output process variable<br>**Step 1: Find Equations**<br>Search for equations representing the output process variable as a function of input variables<br>**if** equation is available **then**<br>    Print equation in general form: output = f(inputs)<br>    Provide a detailed explanation of each component in the equation<br>**end if**<br>**Step 2: Analyze Relationships Between Variables**<br>Use the information from PDF documents<br>**for** all input\_variables **do**<br>    Analyze relationship with output process variable<br>    **if** relationship is identified **then**<br>        Store relationship type (e.g., positive, negative, logarithmic, polynomial, exponential, etc.)<br>    **end if**<br>    Retrieve the influence of each input variable on the output process variable<br>    Retrieve relationship type (e.g., linear, nonlinear, logarithmic, etc.)<br>    **for** all observed relationship **do**<br>        Retrieve relevant tables or text descriptions as evidence<br>    **end for**<br>    Include explanations of process and variables<br>**end for** |

## 2.3 Model Generation and Iterative Model Refinement

Figure 3 illustrates the model generation and iterative model refinement component of our framework. First, the retrieved information is formatted and combined with a prompt for the LLM (Prompt Form 1 in Table 2) to generate initial analytical models (Fig. 3a). This prompt consists of information on parametric relationships extracted by the retrieval section followed by an instruction to transform this information into equations, specifically instructing the LLM not to rely on its general knowledge for creating the equations. Feeding this prompt to a pretrained LLM (GPT-4o-mini here) yields an initial analytical equation of the parametric relationship. A set of 50 candidate equations is generated to account for potential inaccuracies and LLM hallucination during retrieval (Chen et al., 2024), with each equation created independently from scratch. A temperature parameter ranging randomly between 0.3 and 0.8 is applied to encourage diversity and creativity in equation generation (Peeperkorn et al., 2024; Shojaee et al., 2024). Note that till this point only the functional form of the analytical models is created.

A small experimental dataset is used to fit the constant coefficients in the model. The dataset is split into a training set for fitting each model's constants and a validation set for evaluating the model accuracy based the Coefficient of determination ($R^2$) and Mean Squared Error (MSE). Twenty top-performing models are selected and ranked based on their $R^2$ scores. If any of the $R^2$ scores of these top performing models satisfy the success criterion (validation error less than 2%) then the best performing model is selected for use.

If the success criterion is unsatisfied then iterative refinement is conducted (as in Figs. 3b-c) by using the prompt update instruction (Raman et al., 2022) shown in Table 2-Prompt Form 2. This prompt includes the twenty top-performing models, their $R^2$ values, and the initial prompt that also encompasses the information retrieved from the literature. It provides instructions on the significance of previous models based on their $R^2$ values, ensuring that the model generation process prioritizes the most accurate previous models as sources when creating new ones. Additionally, it specifies improvement strategies, such as algebraic manipulation or the introduction of new terms. The prompt update instruction emphasizes generation of models with new functional forms that yield improved $R^2$. The prompt is used by the LLM (GPT-4o-mini) along with the experimental training dataset to generate new analytical models, fit their constants, and continue the iterative model refinement process. In each iteration, the top twenty models are updated if any newly generated models surpass the previous batch in terms of the $R^2$ score. For the testbeds in this work we generate 50 initial models, use 30 experimental points for fitting, select the top 20 models for iterative model refinement. In each refinement iteration, 20 new models are generated.

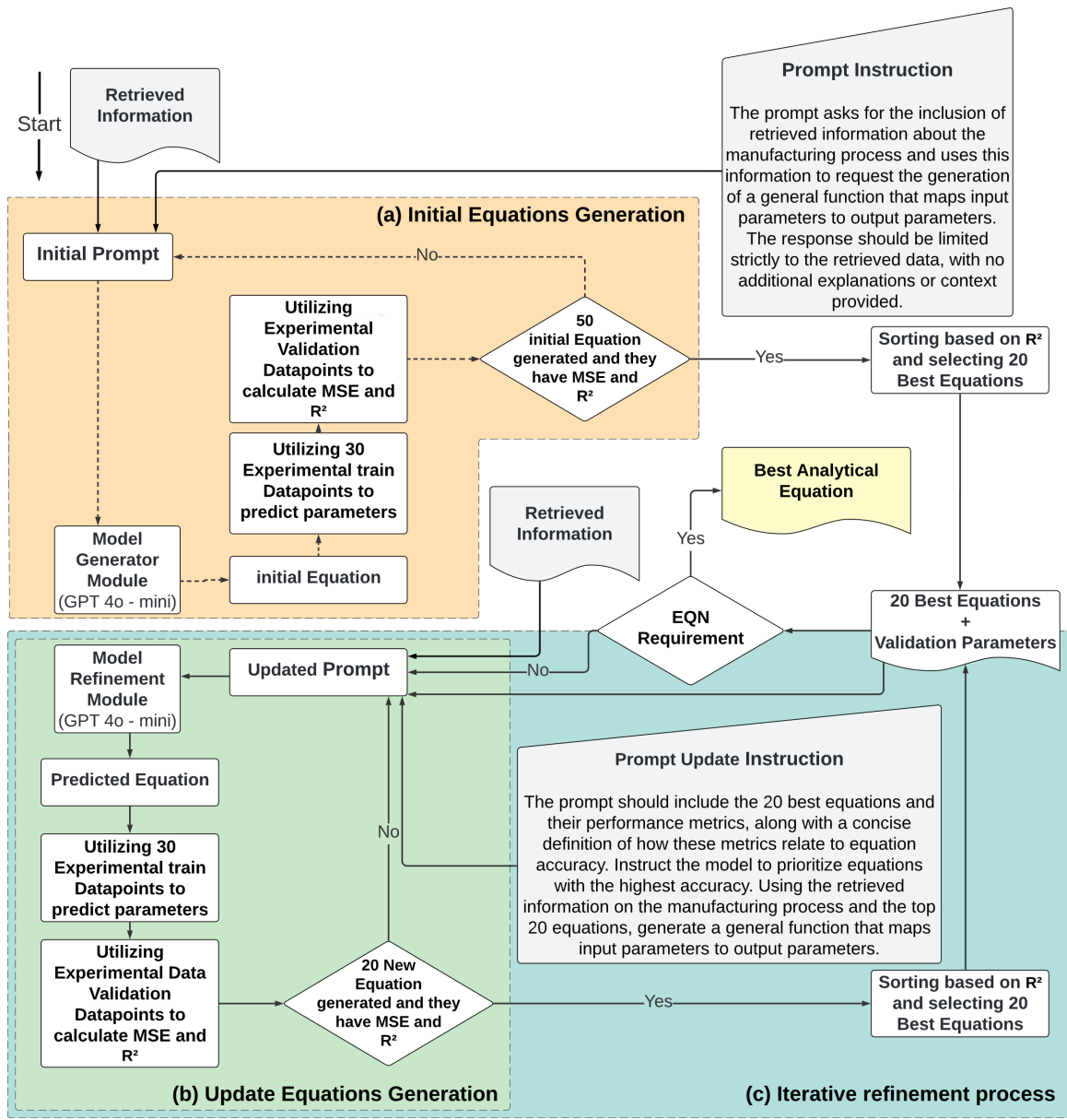

Figure 3: Flow chart for model generation and iterative model refinement component.

Table 2: Prompt forms for iterative model refinement.

**Input:** Retrieved information (response)
**Output:** Analytical model for modeling desired manufacturing process output
Use the following retrieved information:
`response`
**if** an equation is provided (Scenario Eq+ctx) **then**
    Format the equation as specified Python function
**else**
    **if** equation is not available or cannot be formatted as requested (Scenario ctx) **then**
      1. Based on the relationships, generate a Python function to model the output as a function of the inputs.
          • The function format can include various types, such as linear, exponential, logarithmic, or trigonometric functions.
      2. Format the function as:

```
def model(inputs, a0, ...):
    input_1, input_2, input_3, ... = inputs
    output = f(input_1, input_2, input_3, ...)
    return output
```

    **end if**
**end if**
Output only the generated function in the specified format with no additional context or explanation.

---

**Input:** Retrieved information (response), Best Previous Models History (summary)
**Output:** Updated Analytical model for modeling desired manufacturing process output
Use the following retrieved information:
`response`
Use the following model summary for top 20 models:
`summary`
Note:
The R2 scores indicate model fit quality (closer to 1 means better fit).
Models with higher R2 scores and lower MSE are more reliable and should be prioritized.
**Goal:** Generate an improved function with higher R2 scores for modeling the manufacturing process output.
Use previously retrieved information and best equations from model summary
Modify equations by altering operations:
• algebraic manipulation
• Combine terms or introduce new terms
• Modify relationships between input parameters and output parameter
Ensure the modified model fits the specified Python function Format
Output only the generated function in the specified format with no additional context or explanation.

## 3. Testbeds and Data Collection

### 3.1 Overview of Process Testbeds

Three mechanistically different processes were used as the testbeds to evaluate our framework. The first subtractive process is Flow assisted Laser-Induced Plasma Micro-Machining (FLIPMM (Wang et al., 2020), Fig. 4a). In FLIPMM a laser is used to create plasma in a dielectric liquid. The plasma is brought into contact with the workpiece to remove material. The water is continuously flowing to wash out the resulting debris. Our framework was used to model the effects of four process parameters namely laser energy, laser frequency, laser scanning speed, and water speed on four output attributes that include the machined microchannel's width and depth, the material removal rate (MRR) and the heat affected zone (HAZ).

The second additive process is Masked Stereolithography (MSLA (Temiz, 2023), Fig. 4c), which is based on photopolymerization of resin. A vat of resin is exposed to shaped UV light source via an LCD screen mask. The exposed resin forms a deposit of solid polymer layer on the build plate such that the deposit's shape replicates the mask's transparent region. The build plate moves up and the shape of the mask is altered as per the desired slice geometry. A new layer solidifies below the previous layer. Layer deposition is repeated till the full part is printed. Our framework was used to model the effects of layer thickness, exposure time, and build orientation on the printing time and ultimate tensile strength of the printed part.

The third testbed, i.e., Turn-Assisted Deep Cold Rolling (TADCR (Prabhu, Kulkarni, et al., 2020), Fig. 4b), is a deformation-based technique for improving the fatigue life of metallic components. TADCR plastically deforms the surface of a part to induce compressive surface residual stresses. The workpiece rotates on a lathe and a ball roller compresses the surface. Backrest rollers support the workpiece against the roller's forces. Our framework is used to model the effects of rolling force, ball diameter, initial surface roughness, and number of rolling passes on the output attributes of hardness and roughness of the part.

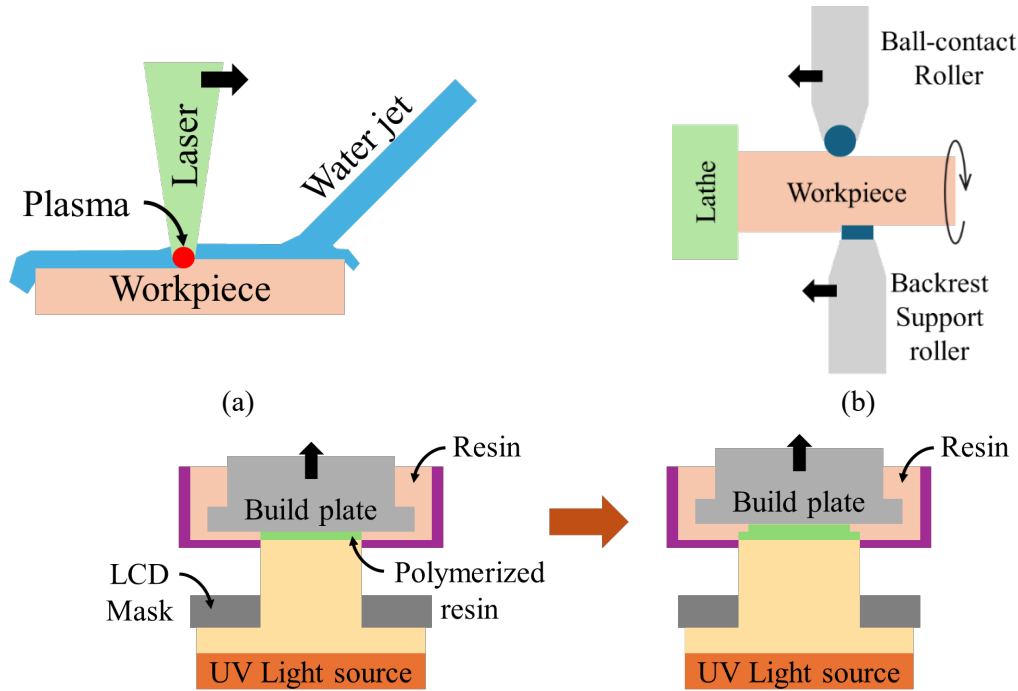

Figure 4: Schematic of testbed processes (a) FLIPMM (b) TADCR and (c) MSLA.

### 3.2 Datasets for Model Generation, Validation, and Testing

The present work used a database of 17 papers related to FLIPMM (Bhandari et al., 2022; Bhandari et al., 2019; Liu et al., 2022; Saxena et al., 2014; Saxena, Malhotra, et al., 2015; Saxena, Wolff, et al., 2015; Tang et al., 2019; Wang et al., 2023; Wang et al., 2022; Wang et al., 2020; Wang et al., 2019; Xie et al., 2020; Zhang et al., 2024; Zhang, Bhandari, et al., 2021; Zhang, Zhang, et al., 2021; Zhang, Liu, et al., 2022; Zhang, Zhang, et al., 2022), 41 papers related to MSLA (Abutaleb et al., 2023; Ahmed et al., 2022; Arslan, 2024; Basson & Bright, 2019; Borra & Neigapula, 2023; de Moraes et al., 2023; Digregorio et al., 2024; Dixon, 2024; Feldmann et al., 2021; Gaikwad et al., 2022; Gür, 2024; Junk & Bär, 2023; Kaufmann et al., 2024; Kricke et al., 2023; Leong et al., 2024; Milovanović et al., 2024; Minin et al., 2021; Mondal et al., 2021; Mondal & Willett, 2022; Montanari et al.; Navarrete-Segado et al., 2021, 2022; Nowacki et al., 2021; Orozco-Osorio et al., 2024; Orzeł & Stecuła, 2022; Ożóg et al., 2022; Paśnikowska-Łukaszuk et al., 2022; Penchev, 2024; Rafalko et al., 2023; Rahman et al., 2024; Sebben et al., 2024; Sharifi et al., 2024; Temiz, 2023, 2024; Torregrosa-Penalva et al., 2022; Valizadeh et al., 2021; Valizadeh et al., 2023; Williams, 2023; P.-J. Yu et al., 2023; Z. Yu et al., 2023; Zuchowicz et al., 2022), and 10 papers related to TADCR (Adıyaman & Aydın, 2024; Kinner-Becker et al., 2021; Luo et al., 2021; Martins et al., 2022; Noronha et al., 2024; Prabhu, Kulkarni, et al., 2020; Prabhu et al., 2014; Prabhu, Prabhu, et al., 2020; Prabhu et al., 2016; Prabhu, 2014), for knowledge retrieval. The papers were chosen from Google Scholar by a non-expert by using the process name as the search keyword and based on a reading of the abstract by the non-expert for potential relevance. This resulted in papers where the relationship to the considered process might traditionally be considered mechanistically incomplete or tenuous, e.g., papers on LIPMM where there is no water flow or papers that also use magnetic fields to manipulate plasma. Note that the automated choice of the papers for knowledge retrieval is relevant to our future work, but is outside the scope of this paper.

Details on the experimental setup and materials for FLIPMM, MSLA, and TADCR can be found in Wang et al. (Wang et al., 2020), Temiz (Temiz, 2023), and Prabhu et al. (Prabhu, Kulkarni, et al., 2020) respectively. The present work uses data from these three papers as the experimental ground truth to evaluate our framework. This synthetic data is generated using equations generated using the Surface Response Methodology in these three papers. Our datasets consist of outputs corresponding to a 4x4x4x4 grid for FLIPMM and TADCR and a 6x6x6 grid for MSLA in the input parameters' space. Cumulatively, the entire dataset consists of 256 points for TADCR and FLIPMM and 216 points for MSLA. These datasets are divided into three subsets: training, validation, and testing.

To evaluate the ability of the generated equations to extrapolate, a capability that is difficult for many data-driven models (Netanyahu et al., 2023), the following sampling strategy was employed. To partition the dataset, a stepwise filtering approach was applied based on input variable value ranges. First, the lowest 75% of values for the first input variable were selected, leaving the top 25% as extrapolation test data. From this selected 75% subset, the second input variable was further filtered by retaining only its lowest 75% of values, while the highest 25% was excluded and added to the test set. This partitioning process continued iteratively across all input variables, and in the end, the remaining data points are randomly assigned to training or validation sets. Therefore, the validation set shares the same value ranges as the training set and is used to evaluate interpolation accuracy, while the test set evaluates extrapolation accuracy. For FLIPMM we used 30 experimental data points for training, 51 for validation, and reserve 175 as the extrapolation test set. For MSLA we used 30 experimental data points for training, 34 for validation, and reserved 152 as the test set. For TADCR we used 30 experimental data points for training, 51 for validation, and reserve 175 as the test set.

The validation dataset enabled evaluation of the interpolation error of generated models and ranking of models during iterative model refinement. The test dataset, for which the input variable's values lie outside the range of inputs for which training and validation are performed, is used to evaluate the extrapolation error of the derived models. The training and validation sets were kept deliberately small to mimic the frequently limited size of experimental data available for training and validation in practical industrial environments. A larger than usual test set was used to ensure robust assessment of how extrapolatable the models derived by our framework are.

## 4. Results

### 4.1. Importance of knowledge type, model refinement, and extrapolative accuracy

Our framework was evaluated for the following scenarios:

**Scenario ctx:** The retrieval process relies entirely on contextual information to identify descriptive relationships between input and output parameters. The retrieval is conducted using Query Form 1 or Step 2 of Query Form 2 when Step 1 cannot extract an explicit equation. This approach is initially used to generate a model without refinement (**Scenario ctx-Initial**), leveraging descriptive parametric relationships. If the generated model does not meet the required accuracy threshold, a refinement component is applied to enhance its accuracy (**Scenario ctx-Refined**).

**Scenario Eq+ctx:** This scenario uses knowledge retrieval to extract analytical models from the literature, if present, in addition to the descriptive relationships extracted in the previous scenario. Retrieval is conducted using Step 1 of Query Form 2. If the extracted knowledge contains a relevant analytical model the function's extracted form is used as a good guess for initial model. Similar to the previous scenario, here we denote **Scenario Eq+ctx – Initial** as the initial model generated by the LLM, and **Scenario Eq+ctx – Refined** as the obtained model after iterative refinement. However, in the presented case studies, we will show that **Eq+ctx – Initial** models have already met the required accuracies; therefore, **Scenario Eq+ctx – Refined** is not activated in these case studies.

Examples of the analytical models generated by our framework are provided. A quantitative comparison of the MSE and $R^2$ on the validation and test sets is performed. In addition, we compare the testing MSE and $R^2$ achieved by our framework to that yielded by direct training of conventional ML models including Gaussian Process Regression (GPR), Support Vector Regression (SVR), and Random Forest Regression (RFR) on the same small experimental training dataset. These comparisons are made separately for each process followed by a summary of overarching conclusions based on the observations.

### 4.1.1. Results for FLIPMM process

Figure 5 shows models derived by our framework for Heat Affected Zone (HAZ) as an exemplar output. Similar information for the other outputs is provided in Figs. A1, A2, and A3 in the Appendix. For Scenario ctx-Initial (Fig. 5a) the derived model does not include interactive effects between inputs and yields a low validation $R^2$ of 0.78 and high validation MSE of 5.2. Model refinement in scenario ctx-Refined (Fig. 5b) increases the model complexity and achieves a drastically higher validation $R^2$ of 0.98 and lower validation MSE of 0.48. Scenario Eq+ctx-Initial (Fig. 5c) achieves an even better validation $R^2$ of 0.99 and validation MSE of 0.0351.

Figure 6 compares the performance of all scenarios over all the outputs for FLIPMM. The Eq+ctx scenario consistently yields the highest performance in extrapolative testing followed by ctx-Refined and then ctx-Initial. For example, for modeling HAZ ctx-refined has a $R^2$ of 0.928 on the extrapolative testing dataset whereas ctx-Initial has $R^2$ of only 0.689. Though the difference in $R^2$ between scenarios ctx-refined and ctx-Initial is not so large for the other model outputs, the $R^2$ for ctx-Refined is always higher than for ctx-Initial.

Table 3 compares the extrapolative testing performance of our framework to traditional ML models. Our framework significantly outperforms SVR, which yields very low and sometimes negative $R^2$ values. Our approach also achieves greater $R^2$ and lower MSE than RFR across the outputs, with the sole exception of the depth output. Further, our framework achieves similar $R^2$ score as GPR for the Width, HAZ and MRR outputs but outperforms GPR for the Depth output by achieving lower MSE and higher $R^2$. Thus, our framework achieves better performance more consistently than traditional ML. Additionally, it outperforms the commonly used symbolic regression model (PySR (Cranmer, 2023)), as detailed in the Appendix (Table A3).

## Scenario ctx:

$(a)$ $\text{HAZ}_{\text{Predicted (Initial)}} = 0.2941 \cdot P + 5.6005 \cdot \dfrac{1}{SS} + 166.305 \cdot \dfrac{1}{WS}$
$$+ 0.0414 \cdot F^2$$

$(b)$ $\text{HAZ}_{\text{Predicted (Refined)}} = 0.0013 \cdot P^2 + 5.5453 \cdot \sqrt{\dfrac{1}{SS}}$
$$+ 235.051 \cdot \dfrac{1}{WS} - 0.1852 \cdot F^{1.5}$$
$$+ 0.131 \cdot (P \cdot F) - 0.6376 \cdot (SS \cdot \sqrt{\dfrac{1}{WS}})$$
$$+ 0.0011 \cdot (P \cdot F^2) - 0.0002 \cdot (P^{1.2} \cdot F \cdot SS)$$
$$- 0.6521 \cdot (F^{0.5} \cdot P^{0.8})$$

## Scenario Eq+ctx:

$(c)$ $\text{HAZ}_{\text{Predicted (Initial)}} = 177.52 - 33.32 \cdot WS - 1.267 \cdot P - 1.282 \cdot F$
$$+ 1.958 \cdot SS + 0.0844 \cdot WS \cdot P - 0.0627 \cdot WS \cdot F$$
$$+ 0.0865 \cdot P \cdot F + 0.1272 \cdot SS^2 + 2.0533 \cdot WS^2$$
$$+ 0.0113 \cdot P^2 + 0.0496 \cdot F^2$$

Figure 5: Analytical models for the Heat Affected Zone for FLIPMM generated using our framework. Notations are Water Speed (WS, m/s), Energy (E, µJ), Frequency (F, kHz), and Scanning Speed (SS, mm/s) as model inputs. The outputs include Channel Width (µm), Channel Depth (µm), Material Removal Rate (MRR, µm³/s), and Heat Affected Zone (HAZ, µm).

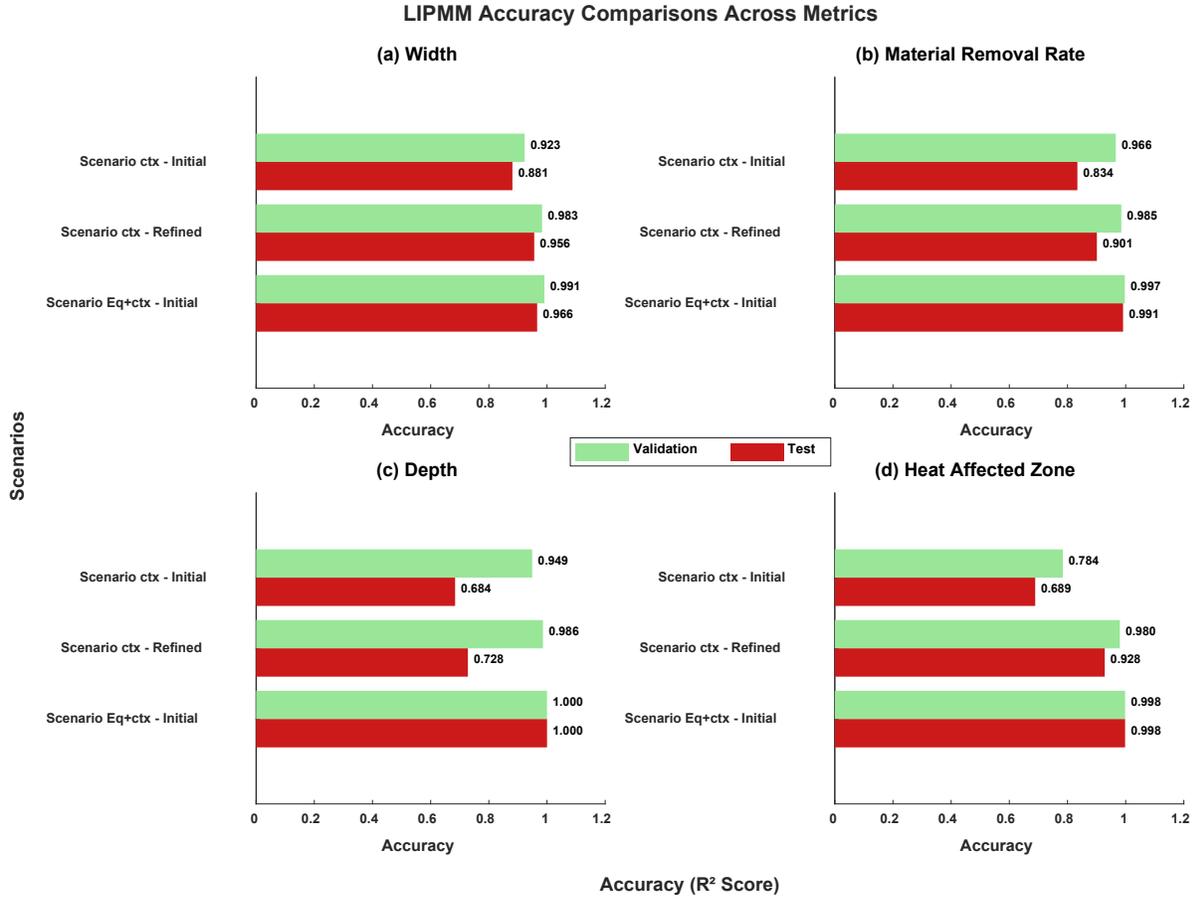

Figure 6: Comparison of models derived by different scenarios of our framework for FLIPMM process.

Table 3: Extrapolative testing for the proposed framework and conventional ML for FLIPMM process.

| Process Output | Model | Test Datapoints | | Process Output | Model | Test Datapoints | |
| | | MSE | $R^2$ Score | | | MSE | $R^2$ Score |
| --- | --- | --- | --- | --- | --- | --- | --- |
| Width | Scenario ctx Refined | $0.601\mu m^2$ | 0.958 | Depth | Scenario ctx Refined | $31.796\mu m^2$ | 0.728 |
| | Scenario Eq+ctx Initial | $0.487\mu m^2$ | 0.966 | | Scenario Eq+ctx Initial | $9.9\text{e-}27\mu m^2$ | 1.0 |
| | GPR | $0.003\mu m^2$ | 1 | | GPR | $55.955\mu m^2$ | 0.522 |
| | SVR | $7.967\mu m^2$ | 0.447 | | SVR | $114.98\mu m^2$ | 0.018 |
| | RFR | $4.406\mu m^2$ | 0.694 | | RFR | $8.768\mu m^2$ | 0.925 |
| Heat Affected | Scenario ctx Refined | $8.118\mu m^6/s^2$ | 0.928 | Material Removal | Scenario ctx Refined | $37.576\mu m^2$ | 0.901 |

| Zone (HAZ) | Scenario Eq+ctx Initial | $0.147\mu m^6/s^2$ | 0.998 | Rate (MRR) | Scenario Eq+ctx Initial | $3.34\mu m^2$ | 0.991 |
|---|---|---|---|---|---|---|---|
| | GPR | $2.875\mu m^6/s^2$ | 0.975 | | GPR | $2.325\mu m^2$ | 0.994 |
| | SVR | $102.221\mu m^6/s^2$ | 0.101 | | SVR | $450.48\mu m^2$ | -0.178 |
| | RFR | $73.305\mu m^6/s^2$ | 0.355 | | RFR | $44.298\mu m^2$ | 0.884 |

### 4.1.2. Results for MSLA process

Figure 7 presents analytical models derived by our framework for Printing Time as the exemplar output. The derived models for the additional outputs are shown in Appendix Figure A4. Figure 8 compares the performance of models derived by our framework for the different scenarios and for all the process outputs. In terms of extrapolation, while scenario Eq+ctx-Initial has the best performance, the performance of ctx-Refined outstrips that of ctx-Initial. Specifically, the testing $R^2$ score for the printing time output is 0.835 for scenario ctx-Refined but only 0.35 for scenario ctx-Initial. Even though the difference between these two scenarios is relatively smaller for the tensile strength output the testing $R^2$ is still higher for ctx-refined.

Table 4 compares the extrapolative testing performance of our framework to traditional ML techniques. SVR and GPR underperform significantly by yielding negative R² scores, indicating that they cannot even qualitatively capture the parametric relationship. Though RFR performs better, our framework still yields higher R² and lower MSE. Thus, our framework outperforms traditional ML for the MSLA process as it does for FLIPMM process.

Scenario ctx:

$$(a)\,\text{Printing\_Time}_{\text{Predicted (Initial)}} = 2.3802 + 14.8337 \cdot \frac{1}{L} - 1.2479 \cdot E$$
$$+ 0.1218 \cdot O + O \cdot \frac{1}{L}$$

$$(b)\,\text{Printing\_Time}_{\text{Predicted (Refined)}} = 59.4384 - 0.9263 \cdot \left(\frac{1}{L}\right)^{1.5} - 1.4472 \cdot E^2$$
$$+ 0.0261 \cdot O^2 + 0.3395 \cdot E \cdot O + 1.7735 \cdot \frac{1}{L} \cdot E$$
$$+ 0.1091 \cdot \frac{1}{L} \cdot O - 0.0667 \cdot E^{1.5} + O \cdot \left(\frac{1}{L}\right)^2$$

Scenario Eq+ctx:

$$(c)\,\text{Printing\_Time}_{\text{Predicted (Initial)}} = -436.00 - 12611.00 \cdot L + 182.00 \cdot E + 17.68 \cdot O$$
$$+ 93828.00 \cdot L^2 - 7.47 \cdot E^2 - 0.0856 \cdot O^2$$
$$- 320.00 \cdot L \cdot E - 95.60 \cdot L \cdot O + 0.027 \cdot E \cdot O$$

Figure 7: Analytical models for the Printing time in MSLA generated using our framework. Notation- Layer Thickness (L, mm), Exposure Time (E, s), and Build Orientation (O, degrees). The output parameters include Ultimate Tensile Strength (UTS, MPa) and Printing Time (Minutes)

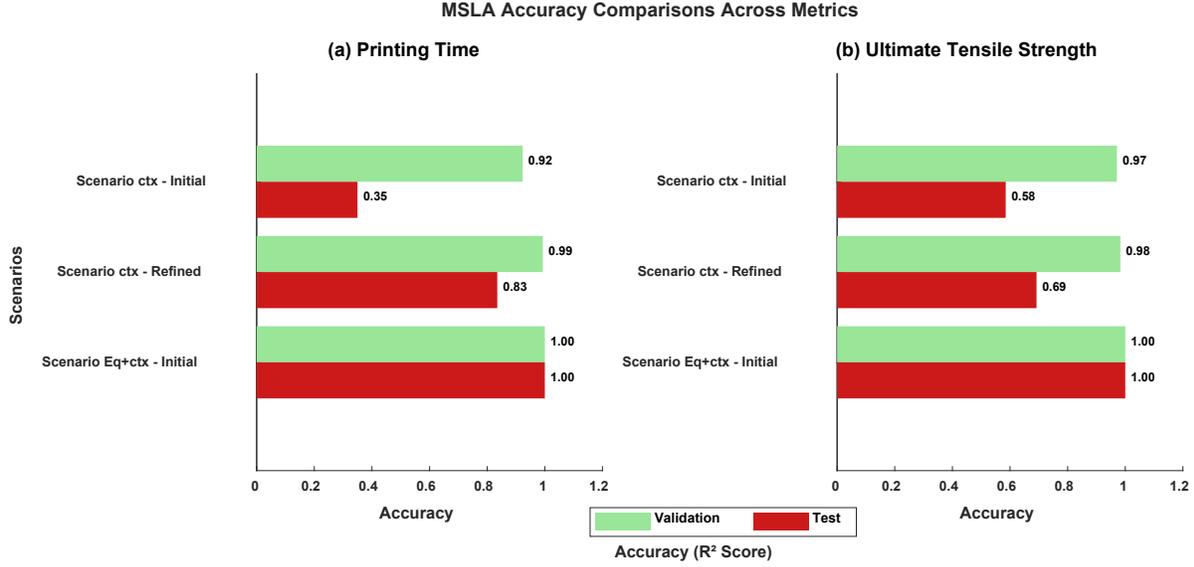

Figure 8: Comparison of models derived by different scenarios of our framework for MSLA process

Table 4: Extrapolative testing for the proposed framework and conventional ML for MSLA process.

| Process Output | Model | Test Datapoints | | Process Output | Model | Test Datapoints | |
|---|---|---|---|---|---|---|---|
| | | MSE | $R^2$ Score | | | MSE | $R^2$ Score |
| Ultimate Tensile Strength (UTS) | Scenario ctx Refined | 11.164MPa² | 0.692 | Printing Time | Scenario ctx Refined | 23393.236 Min² | 0.835 |
| | Scenario Eq+ctx Initial | 1.0e-27MPa² | 1.0 | | Scenario Eq+ctx Initial | 5.5e-25 Min² | 1.0 |
| | GPR | 2413.851MPa² | -65.578 | | GPR | 235634.497 Min² | -0.662 |
| | SVR | 47.081MPa² | -0.299 | | SVR | 151860.637 Min² | -0.071 |
| | RFR | 8.621MPa² | 0.762 | | RFR | 41423.807 Min² | 0.708 |

### 4.1.3. Results for TADCR process

Figure 9 presents models derived by our framework for Roughness as an example. The equations for Surface Hardness are presented in Figure A5 of the Appendix. Figure 10 compares the performance of the models derived by different scenarios of our framework for all the modeled process outputs. Again, scenario Eq+ctx-Initial consistently has the best performance on the test dataset with ctx-Refined next followed by ctx-Initial. For example, the $R^2$ for roughness on the extrapolative dataset is 0.949 for ctx-Refined but only 0.77 for ctx-Initial. While the difference between these two scenarios is relatively smaller for the surface hardness output the corresponding $R^2$ is still higher for ctx-refined. Table 5 compares our framework to conventional ML in an extrapolative context, i.e., on the testing dataset. For this TADCR process, GPR and SVR yield negative R², indicating an inability to qualitatively capture the parametric relationship with the small training dataset that is available. RFR performs better with a positive R² score, but still has substantially lower R² score and higher MSE than our framework. Thus, our framework outperforms traditional ML for TADCR process as well, like it does for the MSLA and FLIPMM processes.

### Scenario ctx:

$(a)$ Roughness$_{\text{Predicted (Initial)}} = 0.9057 - 0.0298 \cdot B - 0.0000755 \cdot R + \dfrac{0.1126}{I}$
$$- 0.1126 \cdot N$$

$(b)$ Roughness$_{\text{Predicted (Refined)}} = 0.6043 - 0.0241 \cdot B^{1.2} - 0.00029 \cdot R^{1.1}$
$$+ \frac{2.0127}{I + 0.01} - 0.0731 \cdot N^{1.3} + 0.000742 \cdot B \cdot R^{0.75}$$
$$- \frac{0.00859 \cdot B^{1.2} \cdot R^{0.5}}{I + 0.01}$$

### Scenario Eq+ctx:

$(c)$ Roughness$_{\text{Predicted (Initial)}} = 1.7845 - 0.1307 \cdot B - 0.00079 \cdot R - 0.1023 \cdot I$
$$- 0.1515 \cdot N + 0.00006 \cdot B \cdot R + 0.00691 \cdot B \cdot N$$
$$+ 0.01104 \cdot B \cdot I - 0.00002 \cdot R \cdot N - 0.00005 \cdot R \cdot I$$
$$- 0.00052 \cdot I \cdot N - 3.30 \times 10^{-9} \cdot N^2$$

Figure 9: Analytical functions for Roughness in TADCR generated using our framework. Notation- Ball Diameter (B, mm), Rolling Force (R, N), Initial surface roughness of the part, (I, μm), and Number of Rolling Passes (N,-), with outputs including Surface Hardness (HV) and Roughness (Ra).

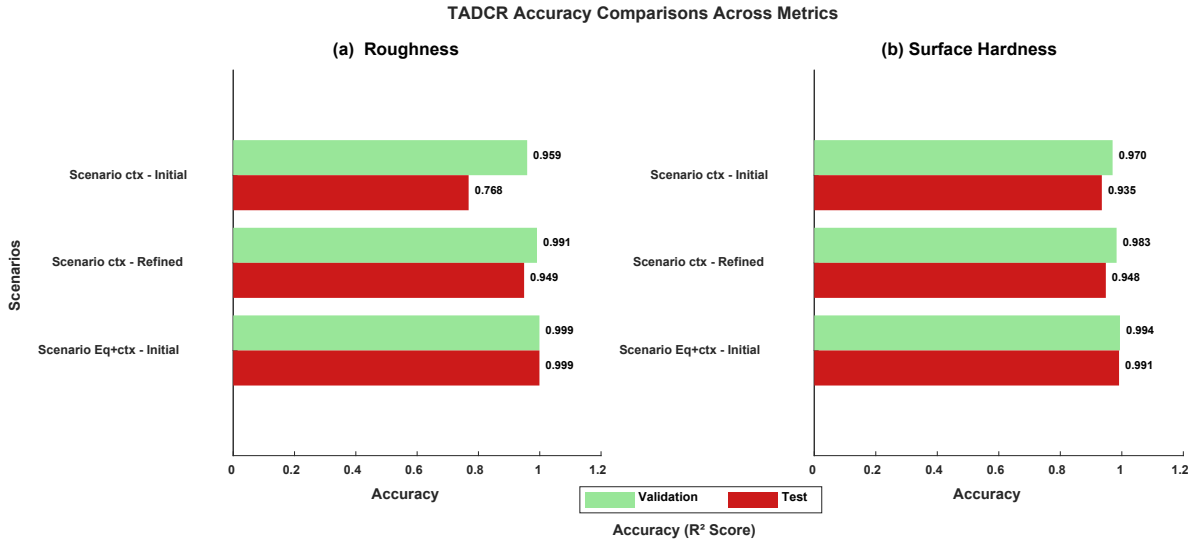

Figure 10: Comparison of models derived by different scenarios of our framework for TADCR process

Table 5: Extrapolative testing for the proposed framework and conventional ML for TADCR process.

| Process Output | Model | Test Datapoints | | Process Output | Model | Test Datapoints | |
|---|---|---|---|---|---|---|---|
| | | MSE | $R^2$ Score | | | MSE | $R^2$ Score |
| Surface Hardness | Scenario ctx Refined | $12.639HV^2$ | 0.948 | Roughness | Scenario ctx Refined | $3.7e\text{-}4Ra^2$ | 0.949 |
| | Scenario Eq+ctx Initial | $2.087HV^2$ | 0.991 | | Scenario Eq+ctx Initial | $1.3e\text{-}17Ra^2$ | 0.999 |
| | GPR | $30980.965HV^2$ | -124.15 | | GPR | $0.068Ra^2$ | -8.152 |
| | SVR | $300.273HV^2$ | -0.213 | | SVR | $0.013Ra^2$ | -0.784 |
| | RFR | $94.14HV^2$ | 0.62 | | RFR | $0.003Ra^2$ | 0.571 |

### 4.1.4. Summary of observations in this section

The above results reveal key insights, as follows

*Question 1: How do the extrapolative ability and data hunger of our approach compare to traditional ML?*

The results show that across the range of processes and input-output combinations examined here our framework yields models with high correlation and low error much more consistently than traditional ML. This advantage is realized on the extrapolative test dataset, for which the traditional ML models often exhibit negative $R^2$ values and very high MSE and are thus unable to capture the parametric relationship. Both our framework and the traditional ML models are trained on the same small experimental dataset. Thus, our framework utilizes experimental data much more efficiently while enabling better predictive capability in extrapolation.

A potential reason is that our framework's ability to combine descriptive relationships and data allows the resulting models to remain grounded in physical reality much more than a traditional regressor, and thus allows better extrapolation across a broader range of input-output combinations and processes. For example, a descriptive statement extracted from the literature such as "microchannel width increases with laser power" provides a form of physical regularization that allows better extrapolation that trying to derive this trend solely from the data. A formal testing of this hypothesis is part of continuing work by the authors.

*Question 2: Is it desirable to retrieve only descriptive relationships or equations and descriptive relationships from the literature?*

The results show that retrieval of equations and descriptive relationships (scenario Eq+ctx) yields the greatest accuracy in extrapolative testing. Thus, extraction of both equations and descriptive relationships is preferable. The authors encourage the broader manufacturing community to provide equations along with descriptions of parametric relationships in their papers since this is necessary for scenario Eq+ctx.

*Question 3: If scenario Eq+ctx is so accurate then what is the value of the model refinement component?*

Papers in the literature may not provide the closed-form equations necessary for retrieval of equations. It is possible that only descriptive relationships, e.g., descriptions of trends in graphs or of finite element analyses, are available. This is especially true for novel processes where work is still nascent. In such a case scenario Eq+ctx is no longer possible. The results show that scenario ctx-Refined, in which retrieval of descriptive relationships is combined with model refinement, is more accurate in extrapolation than scenario ctx-Intial, in which only knowledge retrieval is used without model refinement. Thus, model refinement is critical when the literature does not contain closed-form equations.

### 4.2. Significance of Knowledge Retrieval

This section answers the question *Is knowledge retrieval from the literature necessary for our framework?*

We performed an ablation study in which our framework was used to generate analytical models without using retrieved information from the literature (i.e., without-RAG), thus relying solely on data-driven iterative model refinement. The prompt forms are shown in Table A1 in the Appendix. Table 6 shows the testing $R^2$ for the initial and refined models without RAG and compares it to the testing $R^2$ for the with-RAG cases from the previous sections. Results for the validation set, representing interpolative prediction, are in Table A2 of the Appendix.

Table 6 shows that the models generated without RAG often fail to or struggle to generalize. For instance, for predicting the Heat-Affected Zone (HAZ) in FLIPMM without RAG, the $R^2$ dropped from 0.67 in the initial model to -44.863 after refinement. When RAG was used the $R^2$ increased from initial value of 0.689 to a refined final value of 0.928. For the other outputs in the FLIPMM process the $R^2$ after refinement was greater with RAG than without RAG. In MSLA, incorporating RAG significantly improved $R^2$ for Printing Time. Without RAG, the initial $R^2$ score was -0.296 and only a slight improvement to -0.076 was observed after model refinement. Using RAG yielded a higher initial $R^2$ of 0.35 which improved dramatically to 0.835 after refinement. At the same time, the $R^2$ of the refined with-RAG model for Tensile strength was similar to that of the refined without-RAG model. For both the outputs of the TADCR process, the with-RAG and without-RAG models had similar performance in terms of the $R^2$ of the refined model.

Thus, the value of the RAG-based knowledge retrieval component of our framework lies in ensuring greater stability and consistency in extrapolative predictions beyond the experimental training data. It is possible that the earlier-described physical regularization via inclusion of RAG-extracted descriptive relationships from the literature (see Section 4.1.4) is the reason for this observation as well.

Table 6: Comparison of R² for models generated with and without RAG across the process testbeds.

### FLIPMM Process

| Process Output | Model | Test R² score | | Process Output | Model | Test R² score | |
|---|---|---|---|---|---|---|---|
| | | Without RAG | With RAG | | | Without RAG | With RAG |
| Width | Initial | 0.838 | 0.881 | Depth | Initial | -0.356 | 0.684 |
| | Refined | 0.917 | 0.958 | | Refined | 0.691 | 0.728 |
| Heat Affected Zone (HAZ) | Initial | 0.67 | 0.689 | Material Removal Rate (MRR) | Initial | 0.434 | 0.834 |
| | Refined | -44.863 | 0.928 | | Refined | 0.841 | 0.901 |

### MSLA Process

| Process Output | Model | Test R² score | | Process Output | Model | Test R² score | |
|---|---|---|---|---|---|---|---|
| | | Without RAG | With RAG | | | Without RAG | With RAG |
| Ultimate Tensile Strength (UTS) | Initial | 0.483 | 0.585 | Printing Time | Initial | -0.296 | 0.350 |
| | Refined | 0.769 | 0.692 | | Refined | -0.076 | 0.835 |

### TADCR Process

| Process Output | Model | Test R² score | | Process Output | Model | Test R² score | |
|---|---|---|---|---|---|---|---|
| | | Without RAG | With RAG | | | Without RAG | With RAG |
| Surface Hardness | Initial | 0.904 | 0.935 | Roughness | Initial | 0.784 | 0.768 |
| | Refined | 0.960 | 0.948 | | Refined | 0.949 | 0.949 |

## 5. Conclusion

This paper establishes a novel LLM-based framework for automated modeling of parametric relationships in data-deficient manufacturing processes. The novelty lies in combining RAG-based knowledge retrieval with data-driven iterative model refinement. This framework is evaluated for three manufacturing process testbeds with mechanistically different operational principles.

The results show that our framework goes beyond the state-of-the-art use of LLMs for the same purpose by (i) eliminating the need for human intervention such as interpretation of literature, manual problem specification, and generation of high-performing initial equations; and (ii) simultaneously incorporating experimental data to increase accuracy beyond that possible with just RAG (see discussion in Section 4.1.4). As compared to traditional ML the models derived by our framework have significantly and surprisingly better predictive capability in extrapolation despite the use of a small training dataset (as discussed in Section 4.1.4). The reason is likely a partial physical regularization provided by descriptive relationships extracted by RAG.

The following additional insights are revealed:

1. As discussed in Section 4.1.3, the extraction of both equations and descriptive relationships via RAG is preferable. But the alternative of using descriptive relationships with model refinement is equally important. This is because the literature may not provide closed-form equations, e.g., papers might only textually describe trends observed in graphs or numerical models. At the same time, given the accuracy of models based on extraction of equations and descriptive relationships, the manufacturing community is encouraged to include closed-form parametric equations in papers when possible.

2. RAG-based knowledge retrieval is critical since it provides greater stability in extrapolation, as discussed in Section 4.2. This could be due to physical regularization provided by descriptive relationships extracted from the literature by RAG.

The ability of the developed framework to use minimal experimental data and eliminate subjective human modeling or interpretation will enable accelerated adoption of novel processes for which the underlying physics is not well known (e.g., material removal in FLIPMM) and of existing processes for which part of the physics is not known or is difficult to model (e.g., fatigue life in additive processes). Our future work will explore extension of our framework towards spatiotemporal modeling of material states (e.g., stress) and properties (e.g., surface finish) and incorporate advanced retrieval methods such as hybrid search .

**Acknowledgement**

The authors acknowledge financial support from the National Science Foundation grants # CMMI-2414398 and # CMMI-2336448. KNK also gratefully acknowledges the Pratt & Whitney Institute for Advanced Systems Engineering Fellowship from the University of Connecticut.

**Conflicts of Interests**

The authors declare that they have no known competing funding, financial, employment, or any other non-financial interests that could have appeared to influence the work reported in this paper.

**Data Availability**

Upon acceptance, the data and model for reproducing the results will be made available upon email request.

**References**

Abutaleb, A., Alhamad, M., Al-Orf, M., Alrashed, K., & Esmaeili, S. E. (2023). The design of a dual open-source 3d printer utilizing FDM and MSLA printing technologies. *Int. J. Mechatron. Appl. Mech.*, *2*, 103.

Adıyaman, O., & Aydın, F. (2024). Deep Rolling of Al6061-T6 Material and Performance Evaluation with New Type Designed WNMG Formed Rolling Tool. *Celal Bayar University Journal of Science*, *20*(1), 29-40.

Ahmed, I., Sullivan, K., & Priye, A. (2022). Multi-resin masked stereolithography (MSLA) 3D printing for rapid and inexpensive prototyping of microfluidic chips with integrated functional components. *Biosensors*, *12*(8), 652.

Allison, J., Li, M., Wolverton, C., & Su, X. (2006). Virtual aluminum castings: an industrial application of ICME. *JOM*, *58*, 28-35.

Ampazis, N. (2024). Improving RAG Quality for Large Language Models with Topic-Enhanced Reranking. IFIP International Conference on Artificial Intelligence Applications and Innovations,

Arslan, S. (2024). *A Run Length Encoding Based Variable Byte Compression of Binary Layer Projection Images for Msla and DLP 3D Printers* Middle East Technical University (Turkey)].

Bao, W., Chu, X., Lin, S., & Gao, J. (2015). Experimental investigation on formability and microstructure of AZ31B alloy in electropulse-assisted incremental forming. *Materials & Design*, *87*, 632-639. https://doi.org/https://doi.org/10.1016/j.matdes.2015.08.072

Basson, C., & Bright, G. (2019). Investigation of an Economical Stereolithography 3d Printer for Rapid Prototyping and Mass Production. *COMA'19*, 194.

Bhandari, S., Kang, P., Jeong, J., Cao, J., & Ehmann, K. (2022). Cavitation bubble removal by surfactants in Laser-Induced Plasma Micromachining. *Manufacturing Letters*, *32*, 96-99. https://doi.org/https://doi.org/10.1016/j.mfglet.2022.04.004

Bhandari, S., Murnal, M., Cao, J., & Ehmann, K. (2019). Comparative Experimental Investigation of Micro-channel Fabrication in Ti Alloys by Laser Ablation and Laser-induced Plasma Micro-machining. *Procedia Manufacturing*, *34*, 418-423. https://doi.org/https://doi.org/10.1016/j.promfg.2019.06.186

Borra, N. D., & Neigapula, V. S. N. (2023). Parametric optimization for dimensional correctness of 3D printed part using masked stereolithography: Taguchi method. *Rapid Prototyping Journal*, *29*(1), 166-184.

Buehler, M. J. (2024). MechGPT, a Language-Based Strategy for Mechanics and Materials Modeling That Connects Knowledge Across Scales, Disciplines, and Modalities. *Applied Mechanics Reviews*, *76*(2), 021001.

Chandrasekhar, A., Chan, J., Ogoke, F., Ajenifujah, O., & Farimani, A. B. (2024). AMGPT: a Large Language Model for Contextual Querying in Additive Manufacturing. *arXiv preprint arXiv:2406.00031.*

Chang, Z., Yang, M., & Chen, J. (2021). Experimental investigations on deformation characteristics in microstructure level during incremental forming of AA5052 sheet. *Journal of Materials Processing Technology*, *291*, 117006. https://doi.org/https://doi.org/10.1016/j.jmatprotec.2020.117006

Chen, M., Malhotra, R., & Guo, W. G. (2023). Transfer learning for predictive quality in laser-induced plasma micro-machining. International Manufacturing Science and Engineering Conference,

Chen, X., Wang, L., Wu, W., Tang, Q., & Liu, Y. (2024). Honest AI: Fine-Tuning" Small" Language Models to Say" I Don't Know", and Reducing Hallucination in RAG. *arXiv preprint arXiv:2410.09699.*

Cleeman, J., Agrawala, K., Nastarowicz, E., & Malhotra, R. (2023). Partial-physics-informed multi-fidelity modeling of manufacturing processes. *Journal of Materials Processing Technology*, *320*, 118125.

Cleeman, J., Bogut, A., Mangrolia, B., Ripberger, A., Kate, K., Zou, Q., & Malhotra, R. (2022). Scalable, flexible and resilient parallelization of fused filament fabrication: Breaking endemic tradeoffs in material extrusion additive manufacturing. *Additive Manufacturing*, *56*, 102926.

Cranmer, M. (2023). Interpretable machine learning for science with PySR and SymbolicRegression. jl. *arXiv preprint arXiv:2305.01582.*

de Moraes, N. C., Daakour, R. J. B., Pedão, E. R., Ferreira, V. S., da Silva, R. A. B., Petroni, J. M., & Lucca, B. G. (2023). Electrochemical sensor based on 3D-printed substrate by masked stereolithography (MSLA): a new, cheap, robust and sustainable approach for simple production of analytical platforms. *Microchimica Acta*, *190*(8), 312.

Devaraj, H., Hwang, H.-J., & Malhotra, R. (2020). Understanding the role of nanomorphology on resistance evolution in the hybrid form-fuse process for conformal electronics. *Journal of Manufacturing Processes*, *58*, 1088-1102.

Devaraj, H., & Malhotra, R. (2019). Scalable forming and flash light sintering of polymer-supported interconnects for surface-conformal electronics. *Journal of Manufacturing Science and Engineering*, *141*(4), 041014.

Devaraj, H., Tian, Q., Guo, W., & Malhotra, R. (2021). Multiscale Modeling of Sintering-Driven Conductivity in Large Nanowire Ensembles. *ACS Applied Materials & Interfaces*, *13*(47), 56645-56654.

Digregorio, G., Calderon, A. J., Lebailly, A., & Redouté, J.-M. (2024). Stereolithography LED-LCD 3D Printing with Sub-5 Micron Layer Thickness. *IEEE Sensors Letters*.

Dixon, S. (2024). The Effect of Extended Saltwater Absorption and UV Cure on the Compressive Properties of MSLA 3D-Printed Photopolymer Resin Samples with Printing Variations.

Du, M., Chen, Y., Wang, Z., Nie, L., & Zhang, D. (2024). Large language models for automatic equation discovery of nonlinear dynamics. *Physics of Fluids*, *36*(9).

Eslaminia, A., Jackson, A., Tian, B., Stern, A., Gordon, H., Malhotra, R., Nahrstedt, K., & Shao, C. (2024). FDM-Bench: A Comprehensive Benchmark for Evaluating Large Language Models in Additive Manufacturing Tasks. *arXiv preprint arXiv:2412.09819*.

Feldmann, J., Spiehl, D., & Dörsam, E. (2021). Paper embossing tools: a fast fabrication workflow using image processing and stereolithography additive manufacturing. Advances in Printing and Media Technology: Proceedings of the 47th International Research Conference of iarigai. Athens, Greece,

Fernández-Godino, M. G. (2016). Review of multi-fidelity models. *arXiv preprint arXiv:1609.07196*.

Fuchs, S. L., Praegla, P. M., Cyron, C. J., Wall, W. A., & Meier, C. (2022). A versatile SPH modeling framework for coupled microfluid-powder dynamics in additive manufacturing: binder jetting, material jetting, directed energy deposition and powder bed fusion. *Engineering with Computers*, *38*(6), 4853-4877. https://doi.org/10.1007/s00366-022-01724-4

Gaikwad, S. R., Pawar, N. H., & Sapkal, S. U. (2022). Comparative evaluation of 3D printed components for deviations in dimensional and geometrical features. *Materials Today: Proceedings*, *59*, 297-304.

Glass, M., Rossiello, G., Chowdhury, M. F. M., Naik, A. R., Cai, P., & Gliozzo, A. (2022). Re2G: Retrieve, rerank, generate. *arXiv preprint arXiv:2207.06300*.

Gong, N., Reddy, C. K., Ying, W., & Fu, Y. (2024). Evolutionary Large Language Model for Automated Feature Transformation. *arXiv preprint arXiv:2405.16203*.

Grant, L. O., Higgs, C. F., & Cordero, Z. C. (2023). Sintering mechanics of binder jet 3D printed ceramics treated with a reactive binder. *Journal of the European Ceramic Society*, *43*(6), 2601-2613. https://doi.org/https://doi.org/10.1016/j.jeurceramsoc.2022.12.017

Grayeli, A., Sehgal, A., Costilla-Reyes, O., Cranmer, M., & Chaudhuri, S. (2024). Symbolic regression with a learned concept library. *arXiv preprint arXiv:2409.09359*.

Gür, Y. (2024). Masked stereolithography 3D printing of a brain tissue from an MRI data set. *Alexandria Engineering Journal*, *98*, 302-311.

Huang, K., & Logé, R. E. (2016). A review of dynamic recrystallization phenomena in metallic materials. *Materials & Design*, *111*, 548-574. https://doi.org/https://doi.org/10.1016/j.matdes.2016.09.012

Jadhav, Y., Pak, P., & Farimani, A. B. (2024). Llm-3D print: large language models to monitor and control 3D printing. *arXiv preprint arXiv:2408.14307*.

Jahangir, M. N., Devaraj, H., & Malhotra, R. (2020). On self-limiting rotation and diffusion mechanisms during sintering of silver nanowires. *The Journal of Physical Chemistry C*, *124*(36), 19849-19857.

Junk, S., & Bär, F. (2023). Design guidelines for Additive Manufacturing using Masked Stereolithography mSLA. *Procedia CIRP*, *119*, 1122-1127.

Kaufmann, B. K., Rudolph, M., Pechtl, M., Wildenburg, G., Hayden, O., Clausen-Schaumann, H., & Sudhop, S. (2024). mSLAb–An open-source masked stereolithography (mSLA) bioprinter. *HardwareX*, e00543.

Kennedy, M. C., & O'Hagan, A. (2000). Predicting the output from a complex computer code when fast approximations are available. *Biometrika*, *87*(1), 1-13.

Kinner-Becker, T., Hettig, M., Sölter, J., & Meyer, D. (2021). Analysis of internal material loads and Process Signature Components in deep rolling. *CIRP Journal of Manufacturing Science and Technology*, *35*, 400-409.

Kricke, J. L., Yusnila Khairani, I., Beele, B. B., Shkodich, N., Farle, M., Slabon, A., Doñate-Buendía, C., & Gökce, B. (2023). 4D printing of magneto-responsive polymer structures by masked stereolithography for miniaturised actuators. *Virtual and Physical Prototyping*, *18*(1), e2251017.

Leong, K. M., Sun, A. Y., Quach, M. L., Lin, C. H., Craig, C. A., Guo, F., Robinson, T. R., Chang, M. M., & Olanrewaju, A. O. (2024). Democratizing Access to Microfluidics: Rapid Prototyping of Open Microchannels with Low-Cost LCD 3D Printers. *ACS omega*.

Li, Y., Li, W., Yu, L., Wu, M., Liu, J., Li, W., Wei, S., & Deng, Y. (2024). MLLM-SR: Conversational Symbolic Regression base Multi-Modal Large Language Models. *arXiv preprint arXiv:2406.05410*.

Liu, Y., Guo, H., Wang, H., Zhang, Y., & Zhang, Z. (2022). Comparative Analysis of Bubbles Behavior in Different Liquids by Laser-Induced Plasma Micromachining Single-Crystal Silicon. *Crystals*, *12*(2), 286.

*LlamaParse*. LlamaIndex. https://docs.llamaindex.ai/en/stable/llama_cloud/llama_parse/

Luo, X., Ren, X., Jin, Q., Qu, H., & Hou, H. (2021). Microstructural evolution and surface integrity of ultrasonic surface rolling in Ti6Al4V alloy. *Journal of Materials Research and Technology*, *13*, 1586-1598.

Mansurova, A., Mansurova, A., & Nugumanova, A. (2024). QA-RAG: Exploring LLM Reliance on External Knowledge. *Big Data and Cognitive Computing*, *8*(9), 115.

Mao, Y., Cai, C., Zhang, J., Heng, Y., Feng, K., Cai, D., & Wei, Q. (2023). Effect of sintering temperature on binder jetting additively manufactured stainless steel 316L: densification, microstructure evolution and mechanical properties. *Journal of Materials Research and Technology*, *22*, 2720-2735.

Martins, A. M., Rodrigues, P. C., & Abrão, A. M. (2022). Influence of machining parameters and deep rolling on the fatigue life of AISI 4140 steel. *The International Journal of Advanced Manufacturing Technology*, *121*(9), 6153-6167.

Merler, M., Haitsiukevich, K., Dainese, N., & Marttinen, P. (2024). In-context symbolic regression: Leveraging large language models for function discovery. Proceedings of the 62nd Annual Meeting of the Association for Computational Linguistics (Volume 4: Student Research Workshop),

Milovanović, A., Montanari, M., Golubović, Z., Mărghitaș, M. P., Spagnoli, A., Brighenti, R., & Sedmak, A. (2024). Compressive and flexural mechanical responses of components obtained through mSLA vat photopolymerization technology. *Theoretical and Applied Fracture Mechanics*, *131*, 104406.

Minin, A., Blatov, I., Rodionov, S., & Zubarev, I. (2021). Development of a cell co-cultivation system based on protein magnetic membranes, using a MSLA 3D printer. *Bioprinting*, *23*, e00150.

Mondal, D., Haghpanah, Z., Huxman, C. J., Tanter, S., Sun, D., Gorbet, M., & Willett, T. L. (2021). mSLA-based 3D printing of acrylated epoxidized soybean oil-nano-hydroxyapatite composites for bone repair. *Materials Science and Engineering: C*, *130*, 112456.

Mondal, D., & Willett, T. L. (2022). Enhanced mechanical performance of mSLA-printed biopolymer nanocomposites due to phase functionalization. *Journal of the Mechanical Behavior of Biomedical Materials*, *135*, 105450.

Montanari, M., Milovanovic, A., Marghitas, M. P., Sedmak, A., & Brighenti, R. Compressive and flexural mechanical responses of components obtained through MSLA vat photopolymerization technology. *SIRAMM23*, 43.

Moreira, G. d. S. P., Ak, R., Schifferer, B., Xu, M., Osmulski, R., & Oldridge, E. (2024). Enhancing Q&A Text Retrieval with Ranking Models: Benchmarking, fine-tuning and deploying Rerankers for RAG. *arXiv preprint arXiv:2409.07691*.

Mostafaei, A., Elliott, A. M., Barnes, J. E., Li, F., Tan, W., Cramer, C. L., Nandwana, P., & Chmielus, M. (2021). Binder jet 3D printing—Process parameters, materials, properties, modeling, and challenges. *Progress in Materials Science*, *119*, 100707. https://doi.org/https://doi.org/10.1016/j.pmatsci.2020.100707

Muckley, E. S., Saal, J. E., Meredig, B., Roper, C. S., & Martin, J. H. (2023). Interpretable models for extrapolation in scientific machine learning. *Digital Discovery*, *2*(5), 1425-1435.

Naghavi Khanghah, K., Wang, Z., & Xu, H. (2025). Reconstruction and Generation of Porous Metamaterial Units via Variational Graph Autoencoder and Large Language Model. *Journal of Computing and Information Science in Engineering*, *25*(2).

Navarrete-Segado, P., Tourbin, M., Frances, C., & Grossin, D. (2021). Masked stereolithography of hydroxyapatite bioceramic scaffolds through an integrative approach: From powder tailoring to evaluation of 3D printed parts properties.

Navarrete-Segado, P., Tourbin, M., Frances, C., & Grossin, D. (2022). Masked stereolithography of hydroxyapatite bioceramic scaffolds: From powder tailoring to evaluation of 3D printed parts properties. *Open Ceramics*, *9*, 100235.

Netanyahu, A., Gupta, A., Simchowitz, M., Zhang, K., & Agrawal, P. (2023). Learning to extrapolate: A transductive approach. *arXiv preprint arXiv:2304.14329*.

Noronha, D. J., Sharma, S., Prabhu Parkala, R., Shankar, G., Kumar, N., & Doddapaneni, S. (2024). Deep Rolling Techniques: A Comprehensive Review of Process Parameters and Impacts on the Material Properties of Commercial Steels. *Metals*, *14*(6), 667.

Nowacki, B., Kowol, P., Kozioł, M., Olesik, P., Wieczorek, J., & Wacławiak, K. (2021). Effect of post-process curing and washing time on mechanical properties of mSLA printouts. *Materials*, *14*(17), 4856.

Oddiraju, M., Cleeman, J., Malhotra, R., & Chowdhury, S. (2025). A Differentiable Physics-Informed Machine Learning Approach to Model Laser-Based Micro-Manufacturing Process. *Journal of Manufacturing Science and Engineering*, *147*(5).

OpenAI. *GPT-4o-mini*. OpenAI. https://openai.com/index/gpt-4o-mini-advancing-cost-efficient-intelligence/

OpenAI. *text-embedding-3-small*. OpenAI. https://platform.openai.com/docs/guides/embeddings

Orozco-Osorio, Y. A., Gaita-Anturi, A. V., Ossa-Orozco, C. P., Arias-Acevedo, M., Uribe, D., Paucar, C., Vasquez, A. F., Saldarriaga, W., Ramirez, J. G., & Lopera, A. (2024). Utilization of Additive Manufacturing Techniques for the Development of a Novel Scaffolds with Magnetic Properties for Potential Application in Enhanced Bone Regeneration. *Small*, 2402419.

Orzeł, B., & Stecuła, K. (2022). Comparison of 3D Printout Quality from FDM and MSLA Technology in Unit Production. *Symmetry*, *14*(5), 910.

Ożóg, P., Elsayed, H., Grigolato, L., Savio, G., Kraxner, J., Galusek, D., & Bernardo, E. (2022). Engineering of silicone-based blends for the masked stereolithography of biosilicate/carbon composite scaffolds. *Journal of the European Ceramic Society*, *42*(13), 6192-6198.

Pal, S., Bhattacharya, M., Lee, S.-S., & Chakraborty, C. (2024). A domain-specific next-generation large language model (LLM) or ChatGPT is required for biomedical engineering and research. *Annals of biomedical engineering*, *52*(3), 451-454.

Paśnikowska-Łukaszuk, M., Korulczyk, K., Kapłon, K., Urzędowski, A., & Wlazło-Ćwiklińska, M. (2022). Time Distribution Analysis of 3D Prints with the Use of a Filament and Masked Stereolithography Resin 3D Printer. *Advances in Science and Technology. Research Journal*, *16*(5).

Paudel, B. J., Conover, D., Lee, J.-K., & To, A. C. (2021). A computational framework for modeling distortion during sintering of binder jet printed parts. *Journal of Micromechanics and Molecular Physics*, *06*(04), 95-102. https://doi.org/10.1142/S242491302142008X

Peeperkorn, M., Kouwenhoven, T., Brown, D., & Jordanous, A. (2024). Is temperature the creativity parameter of large language models? *arXiv preprint arXiv:2405.00492*.

Penchev, P. (2024). Ultimate flexural strength and Young's modulus analysis of denture base resins for masked stereolithography 3D printing technology. *Archives of Materials Science and Engineering*, *126*(2).

Prabhu, P., Kulkarni, S., & Sharma, S. (2020). Multi-response optimization of the turn-assisted deep cold rolling process parameters for enhanced surface characteristics and residual stress of AISI 4140 steel shafts. *Journal of Materials Research and Technology*, *9*(5), 11402-11423.

Prabhu, P., Kulkarni, S., & Sharma, S. S. (2014). Turn-assisted deep cold rolling-a cost effective mechanical surface treatment technique for surface hardness enhancement. *Journal of Manufacturing Engineering*, *9*(1), 022-029.

Prabhu, P., Prabhu, D., Sharma, S., & Kulkarni, S. (2020). Surface properties and corrosion behavior of turn-assisted deep-cold-rolled AISI 4140 steel. *Journal of Materials Engineering and Performance*, *29*, 5871-5885.

Prabhu, R., Sharma, S., Jagannath, K., Kumar, K., & Kulkarni, S. (2016). Surface Improvement of Shafts by Turn-Assisted Deep Cold Rolling Process. MATEC Web of Conferences,

Prabhu, R. P. (2014). *Investigation on the effects of process Parameters for fatigue life improvement Using turn-assisted deep cold rolling process* Manipal Institute of Technology, Manipal].

Rafalko, C., Stovall, B. J., Zimudzi, T., & Hickner, M. A. (2023). Tunable Stereolithography Photopolymerization Resin for Molding Water-Soluble Cavities. *ACS Applied Polymer Materials*, *5*(12), 10097-10104.

Rahman, M. S., Phani, A., & Kim, S. (2024). Toward High-Performance Piezoresistive Polymer Derived SiOC Ceramics through Masked Stereolithography 3D Printing with β-Silicon Carbide Nanopowder Reinforcement. *Macromolecular Rapid Communications*, *45*(5), 2300602.

Raman, S. S., Cohen, V., Rosen, E., Idrees, I., Paulius, D., & Tellex, S. (2022). Planning with large language models via corrective re-prompting. NeurIPS 2022 Foundation Models for Decision Making Workshop,

Sadeghi Borujeni, S., Shad, A., Abburi Venkata, K., Günther, N., & Ploshikhin, V. (2022). Numerical simulation of shrinkage and deformation during sintering in metal binder jetting with experimental validation. *Materials & Design*, *216*, 110490. https://doi.org/https://doi.org/10.1016/j.matdes.2022.110490

Saxena, I., Ehmann, K., & Cao, J. (2014). Productivity enhancement in laser induced plasma micromachining by altering the salinity of the dielectric media.

Saxena, I., Malhotra, R., Ehmann, K., & Cao, J. (2015). High-speed fabrication of microchannels using line-based laser induced plasma micromachining. *Journal of Micro-and Nano-Manufacturing*, *3*(2), 021006.

Saxena, I., Wolff, S., & Cao, J. (2015). Unidirectional magnetic field assisted Laser Induced Plasma Micro-Machining. *Manufacturing Letters*, *3*, 1-4. https://doi.org/https://doi.org/10.1016/j.mfglet.2014.09.001

Sebben, M. K., Perottoni, R. d. L., Brandl, C. A., Valentim, M. X. G., Silva, J. R. F. d., Tirloni, B., & Daudt, N. d. F. (2024). Evaluation of graphene addition on 3D resin for MSLA vat polymerization. *Matéria (Rio de Janeiro)*, *29*(3), e20240155.

Sharifi, H., Adib, A., Ahmadi, Z., Gemikonakli, E., & Shahedi Asl, M. (2024). Taguchi optimization of mask stereolithographic 3D printing parameters for tensile strengthening of functionally graded resins. *International Journal on Interactive Design and Manufacturing (IJIDeM)*, 1-12.

Sharlin, S., & Josephson, T. R. (2024). In Context Learning and Reasoning for Symbolic Regression with Large Language Models. *arXiv preprint arXiv:2410.17448*.

Sharma, R., Okada, H., Oba, T., Subramanian, K., Yanai, N., & Pranata, S. (2024). Decoding BACnet Packets: A Large Language Model Approach for Packet Interpretation. *arXiv preprint arXiv:2407.15428*.

Shojaee, P., Meidani, K., Gupta, S., Farimani, A. B., & Reddy, C. K. (2024). Llm-sr: Scientific equation discovery via programming with large language models. *arXiv preprint arXiv:2404.18400*.

Shrivastava, P., & Tandon, P. (2019). Microstructure and texture based analysis of forming behavior and deformation mechanism of AA1050 sheet during Single Point Incremental Forming. *Journal of Materials Processing Technology*, *266*, 292-310. https://doi.org/https://doi.org/10.1016/j.jmatprotec.2018.11.012

Tang, H., Qiu, P., Cao, R., Zhuang, J., & Xu, S. (2019). Repulsive magnetic field–assisted laser-induced plasma micromachining for high-quality microfabrication. *The International Journal of Advanced Manufacturing Technology*, *102*(5), 2223-2229. https://doi.org/10.1007/s00170-019-03370-5

Temiz, A. (2023). The effects of process parameters on tensile characteristics and printing time for masked stereolithography components, analyzed using the response surface method. *Journal of Materials Engineering and Performance*, 1-10.

Temiz, A. (2024). The tensile properties of functionally graded materials in MSLA 3D printing as a function of exposure time. *Journal of Materials and Mechatronics: A*, *5*(1), 49-59.

Tian, J., Hou, J., Wu, Z., Shu, P., Liu, Z., Xiang, Y., Gu, B., Filla, N., Li, Y., & Liu, N. (2024). Assessing Large Language Models in Mechanical Engineering Education: A Study on Mechanics-Focused Conceptual Understanding. *arXiv preprint arXiv:2401.12983*.

Torregrosa-Penalva, G., García-Martínez, H., Ortega-Argüello, Á. E., Rodríguez-Martínez, A., Busqué-Nadal, A., & Ávila-Navarro, E. (2022). Implementation of microwave circuits using stereolithography. *Polymers*, *14*(8), 1612.

Valizadeh, I, Al Aboud, A., Dörsam, E., & Weeger, O. (2021). Tailoring of functionally graded hyperelastic materials via grayscale mask stereolithography 3D printing. *Additive Manufacturing*, *47*, 102108.

Valizadeh, I., Tayyarian, T., & Weeger, O. (2023). Influence of process parameters on geometric and elasto-visco-plastic material properties in vat photopolymerization. *Additive Manufacturing*, *72*, 103641.

Virgolin, M., & Pissis, S. P. (2022). Symbolic regression is NP-hard. *arXiv preprint arXiv:2207.01018*.

Wang, P., Zhang, Z., Hao, B., Liu, D., Zhang, Y., Xue, T., & Zhang, G. (2023). Reducing the taper and the heat-affected zone of CFRP plate by Micro fluid assisted laser induced plasma micro-drilling. *Journal of Manufacturing Processes*, *103*, 226-237.

Wang, P., Zhang, Z., Liu, D., Qiu, W., Zhang, Y., & Zhang, G. (2022). Comparative investigations on machinability and surface integrity of CFRP plate by picosecond laser vs laser induced plasma micro-drilling. *Optics & Laser Technology*, *151*, 108022.

Wang, X., Huang, Y., Wang, X., Xu, B., Feng, J., & Shen, B. (2020). Experimental investigation and optimization of laser induced plasma micromachining using flowing water. *Optics & Laser Technology*, *126*, 106067. https://doi.org/10.1016/j.optlastec.2020.106067

Wang, X., Huang, Y., Xu, B., Xing, Y., & Kang, M. (2019). Comparative assessment of picosecond laser induced plasma micromachining using still and flowing water. *Optics & Laser Technology*, *119*, 105623.

Wei, J., Wang, X., Schuurmans, D., Bosma, M., Xia, F., Chi, E., Le, Q. V., & Zhou, D. (2022). Chain-of-thought prompting elicits reasoning in large language models. *Advances in neural information processing systems*, *35*, 24824-24837.

Williams, N. D. (2023). *INVESTIGATION OF VOLATILE ORGANIC CHEMICALS EMITTED BY MASKED STEREOLITHOGRAPHY PRINTERS* California State University, Sacramento].

Xie, J., Ehmann, K., & Cao, J. (2020). Simulation of Ultrashort Laser Pulse Absorption at the Water–Metal Interface in Laser-Induced Plasma Micromachining. *Journal of Micro-and Nano-Manufacturing*, *8*(4), 041009.

Xu, L., Mohaddes, D., & Wang, Y. LLM Agent for Fire Dynamics Simulations. Neurips 2024 Workshop Foundation Models for Science: Progress, Opportunities, and Challenges,

Yu, P.-J., Tseng, T.-C., Wang, Y.-H., Chang, Y.-C., & Yang, S.-H. (2023). Masked Stereolithography 3D-printed Terahertz Diffractive Lens. 2023 48th International Conference on Infrared, Millimeter, and Terahertz Waves (IRMMW-THz),

Yu, Z., Li, X., Zuo, T., Wang, Q., Wang, H., & Liu, Z. (2023). High-accuracy DLP 3D printing of closed microfluidic channels based on a mask option strategy. *The International Journal of Advanced Manufacturing Technology*, *127*(7), 4001-4012.

Zhang, H., Zhang, R., Gao, L., Yang, Z., Mao, Y., Zhao, N., Lu, J., & Wang, X. (2024). Laser-induced plasma micromachining on surfaces parallel to the incident laser in different solutions. *Optics express*, *32*(10), 16970-16982.

Zhang, Y., Bhandari, S., Xie, J., Zhang, G., Zhang, Z., & Ehmann, K. (2021). Investigation on the evolution and distribution of plasma in magnetic field assisted laser-induced plasma micro-machining. *Journal of Manufacturing Processes*, *71*, 197-211. https://doi.org/https://doi.org/10.1016/j.jmapro.2021.09.017

Zhang, Y., Zhang, Z., Zhang, Y., Liu, D., Wu, J., Huang, Y., & Zhang, G. (2021). Study on machining characteristics of magnetically controlled laser induced plasma micro-machining single-crystal silicon. *Journal of Advanced Research*, *30*, 39-51. https://doi.org/https://doi.org/10.1016/j.jare.2020.12.005

Zhang, Z., Liu, D., Zhang, Y., Xue, T., Huang, Y., & Zhang, G. (2022). Fabrication and droplet impact performance of superhydrophobic Ti6Al4V surface by laser induced plasma micro-machining. *Applied Surface Science*, *605*, 154661. https://doi.org/https://doi.org/10.1016/j.apsusc.2022.154661

Zhang, Z., Zhang, Y., Liu, D., Zhang, Y., Zhao, J., & Zhang, G. (2022). Bubble behavior and its effect on surface integrity in laser-induced plasma micro-machining silicon wafer. *Journal of Manufacturing Science and Engineering*, *144*(9), 091008.

Zheng, C., Liu, Z., Xie, E., Li, Z., & Li, Y. (2023). Progressive-hint prompting improves reasoning in large language models. *arXiv preprint arXiv:2304.09797*.

Zhu, Z., Xue, Y., Chen, X., Zhou, D., Tang, J., Schuurmans, D., & Dai, H. (2023). Large language models can learn rules. *arXiv preprint arXiv:2310.07064*.

Zuchowicz, N. C., Belgodere, J. A., Liu, Y., Semmes, I., Monroe, W. T., & Tiersch, T. R. (2022). Low-cost resin 3-D printing for rapid prototyping of microdevices: Opportunities for supporting aquatic germplasm repositories. *Fishes*, *7*(1), 49.

# Appendix

## A.1 Retrieval-Based Process Modeling using RAG

### Case 1: Flowing Water Laser-Induced Plasma Micro-Machining (FLIPMM)

## Scenario ctx:

$$\text{Width}_{\text{Predicted (Initial)}} = 0.0179 \cdot P \cdot F - \frac{4.0102}{SS} - \frac{60.6551}{WS} + 7.6367$$

Performance Metrics (Initial):
Validation MSE:  0.5609
Validation $R^2$ :  0.9235
Test MSE:  1.7019
Test $R^2$ :  0.8819

$$\begin{aligned}
\text{Width}_{\text{Predicted (Refined)}} = {} & -0.0016 \cdot P^{1.6} \cdot F^{1.3} + 1.8098 \cdot \sqrt{\frac{1}{SS}} \\
& + 0.0777 \cdot WS^{1.7} + 0.0253 \cdot P \cdot F^{1.2} \\
& - 51.081 \cdot \sqrt{\frac{1}{WS}} - 0.0144 \cdot P^{1.5} \cdot \frac{1}{SS} \\
& - 0.0500 \cdot F \cdot WS + 0.0315 \cdot F^{0.5} \cdot P^{1.2} \\
& + 2.3703 \cdot \sqrt{SS}
\end{aligned}$$

Performance Metrics (Refined):
Validation MSE:  0.1229
Validation $R^2$ :  0.9833
Test MSE:  0.6016
Test $R^2$ :  0.9582

## Scenario Eq+ctx:

$$\begin{aligned}
\text{Width}_{\text{Predicted (Initial)}} = {} & 34.12 - 1.8203 \cdot WS - 0.0304 \cdot P - 0.0316 \cdot F \\
& - 1.8266 \cdot SS + 0.0822 \cdot WS \cdot SS + 0.0241 \cdot P \cdot F \\
& + 0.0178 \cdot P \cdot SS - 0.0154 \cdot F \cdot SS - 0.0496 \cdot SS^2 \\
& + 0.0033 \cdot F^2
\end{aligned}$$

Performance Metrics (Initial):
Validation MSE:  0.0613
Validation $R^2$ :  0.9916
Test MSE:  0.4878
Test $R^2$ :  0.9661

Figure A1: The presented analytical functions for Width, generated by the LLM in two scenarios (ctx and Eq+ctx).

## Scenario ctx:

$$\text{Depth}_{\text{Predicted (Initial)}} = 0.3660 \cdot P - \frac{21.9713}{SS} + 2.8261 \cdot WS$$
$$+ 1.1399 \cdot F - 32.6477$$

Performance Metrics (Initial):
Validation MSE:  6.2057
Validation $R^2$:  0.9497
Test MSE:  36.9380
Test $R^2$:  0.6847

$$\text{Depth}_{\text{Predicted (Refined)}} = -0.0008 \cdot P^{1.4} \cdot F^{1.3} + 0.0678 \cdot WS^{1.5} \cdot F$$
$$- \frac{9.3292}{SS + 0.1} - 0.0048 \cdot F^2 \cdot P^{0.5} + 0.0512 \cdot P \cdot WS$$
$$- 3.1572 \cdot SS^{0.6} + \frac{0.0848 \cdot P \cdot F}{1 + SS} - 0.2729 \cdot SS^{1.2}$$

Performance Metrics (Refined):
Validation MSE:  1.6136
Validation $R^2$:  0.9869
Test MSE:  31.7967
Test $R^2$:  0.7286

## Scenario Eq+ctx:

$$\text{Depth}_{\text{Predicted (Initial)}} = -97.7573 + 25.4958 \cdot WS + 1.8038 \cdot P + 2.0115 \cdot F$$
$$- 4.8792 \cdot SS - 0.0600 \cdot WS \cdot P + 0.1250 \cdot WS \cdot F$$
$$- 0.2125 \cdot WS \cdot SS - 0.0131 \cdot P \cdot F - 0.0225 \cdot P \cdot SS$$
$$- 0.1469 \cdot F \cdot SS - 1.6500 \cdot WS^2 - 0.0235 \cdot P^2$$
$$- 0.0563 \cdot F^2 + 0.5656 \cdot SS^2$$

Performance Metrics (Initial):
Validation MSE:  $4.5296 \times 10^{-28}$
Validation $R^2$:  1.0
Test MSE:  $2.5455 \times 10^{-27}$
Test $R^2$:  1.0

Figure A2: The presented analytical functions for Depth, generated by the LLM in two scenarios (ctx and Eq+ctx).

## Scenario ctx:

$$\text{MRR}_{\text{Predicted (Initial)}} = -73.0040 + 1.7243 \cdot P + 4.8351 \cdot WS$$
$$- 0.4323 \cdot (5 - SS)^2 + 19.0991 \cdot \ln(F)$$

Performance Metrics (Initial):
Validation MSE:  10.4521
Validation $R^2$:  0.9668
Test MSE:  63.4355
Test $R^2$:  0.8342

$$\text{MRR}_{\text{Predicted (Refined)}} = -86.9738 + 4.1103 \cdot P + 4.5843 \cdot WS$$
$$- 0.4714 \cdot (5 - SS)^2 + 13.0671 \cdot \ln(F)$$
$$- 0.0707 \cdot P^2 + 0.1648 \cdot (WS \cdot F)$$

Performance Metrics (Refined):
Validation MSE:  4.4092
Validation $R^2$:  0.9860
Test MSE:  37.5760
Test $R^2$:  0.9018

## Scenario Eq+ctx:

$$\text{MRR}_{\text{Predicted}} = -245.9066 + 45.5663 \cdot WS + 6.1778 \cdot P + 3.3894 \cdot F$$
$$+ 3.7635 \cdot SS - 0.3785 \cdot P \cdot WS + 0.0446 \cdot P \cdot F$$
$$+ 0.2496 \cdot WS \cdot F - 2.6399 \cdot WS^2 - 0.0672 \cdot P^2$$
$$- 0.1750 \cdot F^2 - 0.2965 \cdot SS^2$$

Performance Metrics (Predicted):
Validation MSE:  0.8768
Validation $R^2$:  0.9972
Test MSE:  3.3409
Test $R^2$:  0.9913

Figure A3: The presented analytical functions for Material Removal Rate (MRR), generated by the LLM in two scenarios (ctx and Eq+ctx).

**Case 2: Masked Stereolithography**

## Scenario ctx:

$$\text{UTS}_{\text{Predicted (Initial)}} = 50.2946 + 8.1617 \cdot \ln(E) - 155.9657 \cdot L$$
$$- 3.4706 \cdot \sin^2(\text{radians}(O))$$

Performance Metrics (Initial):
Validation MSE:   0.9159
Validation $R^2$:   0.9713
Test MSE:   15.0161
Test $R^2$:   0.5858

$$\text{UTS}_{\text{Predicted (Refined)}} = 43.6926 - 39.6988 \cdot \sqrt{L} + 25.9001 \cdot \ln(E+1)$$
$$- 10.8882 \cdot \sqrt{E} - 1.5444 \cdot \sin(\text{radians}(O))$$
$$\cdot \left(1 - \frac{(O-60)^2}{900}\right) + 0.0006 \cdot (O-45)^2$$
$$- 592.7573 \cdot \frac{L}{10}$$

Performance Metrics (Refined):
Validation MSE:   0.5196
Validation $R^2$:   0.9837
Test MSE:   11.1642
Test $R^2$:   0.6921

## Scenario Eq+ctx:

$$\text{UTS}_{\text{Predicted (Initial)}} = 57.41 - 271.6 \cdot L + 2.601 \cdot E - 0.1371 \cdot O$$
$$+ 637.0 \cdot L^2 - 0.1282 \cdot E^2 - 0.000468 \cdot O^2$$
$$+ 3.63 \cdot L \cdot E + 0.616 \cdot L \cdot O + 0.00959 \cdot E \cdot O$$

Performance Metrics (Initial):
Validation MSE:   $1.9155 \times 10^{-28}$
Validation $R^2$:   1.0
Test MSE:   $1.0971 \times 10^{-27}$
Test $R^2$:   1.0

Figure A4: The presented analytical functions for Ultimate Tensile Strength, generated by the LLM in two scenarios (ctx and Eq+ctx).

**Case 3: Turn-Assisted Deep Cold Rolling**

## Scenario ctx:

$$\text{SurfaceHardness}_{\text{Predicted (Initial)}} = 4.9965 \cdot B + 0.0671 \cdot R - 2.8784 \cdot I + 7.5271 \cdot N$$
$$+ 203.8603$$

Performance Metrics (Initial):
Validation MSE:   3.3015
Validation $R^2$:   0.9707
Test MSE:   15.9102
Test $R^2$:   0.9357

$$\text{SurfaceHardness}_{\text{Predicted (Refined)}} = 229.7337 + 0.0972 \cdot B^2 + 0.0000066 \cdot R^2$$
$$- 0.2488 \cdot I^2 + 7.1043 \cdot N + 0.0081 \cdot B \cdot R$$
$$+ 0.0081 \cdot B \cdot N - 0.0081 \cdot I \cdot N$$

Performance Metrics (Refined):
Validation MSE:   1.8453
Validation $R^2$:   0.9836
Test MSE:   12.6390
Test $R^2$:   0.9489

## Scenario Eq+ctx:

$$\text{SurfaceHardness}_{\text{Predicted (Initial)}} = 164.6055 + 7.8401 \cdot B + 0.0293 \cdot R - 7.8315 \cdot I$$
$$+ 15.8052 \cdot N + 0.0069 \cdot B \cdot R - 1.1228 \cdot B \cdot I$$
$$+ 0.3295 \cdot B \cdot N - 0.0084 \cdot R \cdot N + 1.3442 \cdot I \cdot N$$

Performance Metrics (Initial):
Validation MSE:   0.6352
Validation $R^2$:   0.9944
Test MSE:   2.0877
Test $R^2$:   0.9916

Figure A5: The presented analytical functions for Surface Hardness, generated by the LLM in two scenarios (ctx and Eq+ctx).

## A.2 Model Generation and Iterative Refinement Without RAG

Table A1: Prompt Forms for Analytical Model Generation. The first prompt generate initial equations, while the second focuses on iterative refinement to improve model accuracy.

| **Prompt Form S 1** Generate Initial General Form Function | **Prompt Form S 2** Generate Improved General Form Function |
|---|---|
| **Input:** No Input<br>**Output:** Analytical model for modeling the output Based on inputs<br><br>1. Generate a Python function to model the output as a function of the inputs.<br>   • The function format can include various types, such as linear, exponential, logarithmic, or trigonometric functions.<br>2. Format the function as:<br>     `def model(inputs, a0, ...):`<br>        `x1, x2, x3, ... = inputs`<br>        `y = f(x1, x2, x3, ...)`<br>        `return y`<br><br>Output only the generated function in the specified format with no additional context or explanation. | **Input:** Best Previous Models History (summary)<br>**Output:** Updated Analytical model for modeling the output Based on inputs<br>Use the following model summary for top 20 models:<br>`summary`<br>Note:<br>The R2 scores indicate model fit quality (closer to 1 better fit).<br>Models with higher R2 scores and lower MSE are more reliable and should be prioritized.<br>**Goal:** Generate an improved function with higher R2 scores for modeling the output.<br>Use best equations from model summary<br>Modify equations by altering operations:<br>   • algebraic manipulation<br>   • Combine terms or introduce new terms<br>   • Modify relationships between input parameters and output parameter<br>Ensure the modified function fits the specified Python function Format<br>Output only the generated function in the specified format with no additional context or explanation. |

## A.3 Process Modeling Without RAG

# Without RAG:

$$\text{HAZ}_{\text{Predicted (Initial)}} = 64.0088 - 4.6438 \cdot WS + 0.3399 \cdot P$$
$$+ 0.0436 \cdot F^2 - 4.4275 \cdot \ln(SS + 1)$$

Performance Metrics (Initial):

| | |
|---|---|
| Validation MSE: | 5.2052 |
| Validation $R^2$ : | 0.7845 |
| Test MSE: | 37.5044 |
| Test $R^2$ : | 0.6701 |

$$\text{HAZ}_{\text{Predicted (Refined)}} = 83.3152 - 6.4628 \cdot \ln(WS + 1) \cdot \sqrt{P + 1}$$
$$- 0.0011 \cdot WS^2 \cdot \exp(0.4 \cdot F) - 0.6641 \cdot \sin(P) \cdot \cos(SS)$$
$$+ 0.0228 \cdot WS \cdot P \cdot \ln(F + 2)$$
$$+ 0.3571 \cdot \sqrt{WS + 2} \cdot (P + 1)$$
$$- 3.5734 \times 10^{-7} \cdot \exp(F) \cdot \sqrt{SS + 1}$$
$$- 0.0013 \cdot P^2 \cdot F - 2.0963 \cdot WS \cdot \sin(F)$$
$$- 0.7273 \cdot SS \cdot \ln(WS + P + 1)$$
$$+ 0.2876 \cdot \sqrt{P + 1} \cdot (F + SS)$$
$$+ 0.2536 \cdot WS \cdot F \cdot \cos(P)$$

Performance Metrics (Refined):

| | |
|---|---|
| Validation MSE: | 0.4347 |
| Validation $R^2$ : | 0.9820 |
| Test MSE: | 5214.3507 |
| Test $R^2$ : | −44.8637 |

Figure A6: Analytical functions along with performance metrics for Heat Affected Zone (HAZ) without RAG generated using the proposed framework for two scenarios (ctx and Eq+ctx).

## Without RAG:

$$\text{Width}_{\text{Predicted (Initial)}} = 27.7108 - 1.3239 \cdot WS + 0.2271 \cdot P$$
$$+ 0.2903 \cdot F - 2.8935 \cdot \ln(SS + 1)$$

Performance Metrics (Initial):
Validation MSE:   0.4671
Validation $R^2$ :   0.9363
Test MSE:   2.3234
Test $R^2$ :   0.8387

$$\text{Width}_{\text{Predicted (Refined)}} = 41.1539 - 9.3951 \cdot \sqrt{WS} - 0.0089 \cdot P^2 \cdot \ln(F + 1)$$
$$- 0.0058 \cdot WS \cdot \sin(P) + 4.6077 \cdot \sqrt{F + 1}$$
$$+ 0.4574 \cdot P \cdot \sqrt{F} - 0.0007 \cdot WS \cdot P \cdot SS$$
$$- 0.3384 \cdot \cos(WS) \cdot SS$$
$$- 2.8207 \cdot \sqrt{F} \cdot \ln(P + SS + 1)$$

Performance Metrics (Refined):
Validation MSE:   0.1340
Validation $R^2$ :   0.9817
Test MSE:   1.1948
Test $R^2$ :   0.9171

Figure A7: Analytical functions along with performance metrics for Width without RAG generated using the proposed framework for two scenarios (ctx and Eq+ctx).

## Without RAG:

$$\text{Depth}_{\text{Predicted (Initial)}} = -22.1112 + 3.4508 \cdot WS + 0.5373 \cdot P$$
$$+ 1.1842 \cdot F - 0.4422 \cdot SS^2$$

Performance Metrics (Initial):
Validation MSE:   18.6169
Validation $R^2$ :   0.8491
Test MSE:   158.8852
Test $R^2$ :   −0.3562

$$\text{Depth}_{\text{Predicted (Refined)}} = -9.7354 - 0.2190 \cdot \ln(WS + 1) \cdot P^{1.5}$$
$$- 0.0105 \cdot F^2 \cdot \exp(0.2 \cdot SS) - 0.0471 \cdot \sqrt{P} \cdot WS \cdot F$$
$$- 2.8237 \cdot WS \cdot \sqrt{SS + 1}$$
$$+ 0.6233 \cdot \sin(P) \cdot \ln(F + 1) \cdot SS$$
$$+ 1.5838 \cdot \ln(WS + 1) \cdot P + 0.5943 \cdot (WS^2 + F \cdot \sqrt{P})$$
$$+ 0.8354 \cdot SS^{1.5} + 13.0850 \cdot \frac{WS \cdot \cos(F)}{P + 1}$$

Performance Metrics (Refined):
Validation MSE:   1.7623
Validation $R^2$ :   0.9857
Test MSE:   36.1951
Test $R^2$ :   0.6911

Figure A8: Analytical functions along with performance metrics for Depth without RAG generated using the proposed framework for two scenarios (ctx and Eq+ctx).

## Without RAG:

$$\text{MRR}_{\text{Predicted (Initial)}} = -71.4979 + 5.4462 \cdot WS + 1.5872 \cdot P$$
$$+ 3.3503 \cdot F + 0.1223 \cdot SS^2$$

Performance Metrics (Initial):
Validation MSE:     19.5084
Validation $R^2$ :   0.9380
Test MSE:           216.4138
Test $R^2$ :         0.4342

$$\text{MRR}_{\text{Predicted (Refined)}} = -184.1245 + 0.7465 \cdot WS^2 + 52.8412 \cdot \ln(P+1)$$
$$+ 15.2754 \cdot \sqrt{F} - 2.7892 \cdot \sin(SS)$$
$$- 0.2046 \cdot WS \cdot P + 0.3580 \cdot F^{0.5} \cdot SS$$

Performance Metrics (Refined):
Validation MSE:     3.7985
Validation $R^2$ :   0.9879
Test MSE:           60.5148
Test $R^2$ :         0.8418

Figure A9: Analytical functions along with performance metrics for Material Removal Rate without RAG generated using the proposed framework for two scenarios (ctx and Eq+ctx).

## Without RAG:

$$\text{Printing Time}_{\text{Predicted (Initial)}} = 128.8173 - 8131.9048 \cdot L + 46.2947 \cdot E$$
$$+ 8.0784 \cdot O$$

Performance Metrics (Initial):
Validation MSE:     14237.2398
Validation $R^2$ :   0.8994
Test MSE:           183882.8652
Test $R^2$ :         −0.2967

$$\text{Printing Time}_{\text{Predicted (Refined)}} = 396.1453 - 10942.1783 \cdot \ln(L+1) + 4.9544 \cdot E^2$$
$$+ 1.2754 \cdot O^{1.5} - 54.6442 \cdot L \cdot E \cdot \ln(O+1)$$
$$+ 30570.3912 \cdot L^2 \cdot \sqrt{E+1} - 27.8094$$
$$\cdot \exp(-0.1 \cdot O) \cdot (L+E) \cdot \cos(L)$$
$$+ 25.4484 \cdot \sqrt{L \cdot E} \cdot \ln(O+1) \cdot \sin(E)$$
$$- 0.4224 \cdot L \cdot O \cdot \sqrt{E} \cdot (O+1)$$
$$- 730.0891 \cdot \ln(L+2) \cdot \sin(L) \cdot E$$

Performance Metrics (Refined):
Validation MSE:     1625.7470
Validation $R^2$ :   0.9885
Test MSE:           152719.4531
Test $R^2$ :         −0.0770

Figure A10: Analytical functions along with performance metrics for Printing Time without RAG generated using the proposed framework for two scenarios (ctx and Eq+ctx).

## Without RAG:

$$\text{UTS}_{\text{Predicted (Initial)}} = 58.6287 - 155.5587 \cdot L + 1.0634 \cdot E$$
$$- 0.0439 \cdot O^2$$

Performance Metrics (Initial):
Validation MSE:     1.1797
Validation $R^2$ :   0.9631
Test MSE:           18.7122
Test $R^2$ :         0.4839

$$\text{UTS}_{\text{Predicted (Refined)}} = -481.6610 - 773.5148 \cdot \ln(L+1) + 5.2239 \cdot \sqrt{E}$$
$$- 0.00096 \cdot O^2 + 1.7079 \cdot (L \cdot E)$$
$$+ 535.1203 \cdot \exp(L) + 0.7365 \cdot (L \cdot \sqrt{O})$$

Performance Metrics (Refined):
Validation MSE:     0.5849
Validation $R^2$ :   0.9817
Test MSE:           8.3445
Test $R^2$ :         0.7698

Figure A2: Analytical functions along with performance metrics for Ultimate Tensile Strength (UTS) without RAG generated using the proposed framework for two scenarios (ctx and Eq+ctx).

## Without RAG:

$$\text{Roughness}_{\text{Predicted (Initial)}} = 1.0273 - 0.0298 \cdot B - 0.0000794 \cdot R$$
$$+ 0.0000385 \cdot I^2 - 0.2999 \cdot \ln(N + 1)$$

Performance Metrics (Initial):
Validation MSE:   0.0002371
Validation $R^2$:   0.9542
Test MSE:   0.0016042
Test $R^2$:   0.7850

$$\text{Roughness}_{\text{Predicted (Refined)}} = 2.8180 - 0.5716 \cdot \sqrt{B} - 0.1556 \cdot \ln(R + 1)$$
$$- 0.0070 \cdot I^2 - 0.0352 \cdot N^2$$
$$+ 0.0000134 \cdot (B \cdot R) - 0.01197 \cdot (I \cdot \sin(N))$$
$$+ 0.0020 \cdot (B \cdot I \cdot \ln(R + 1))$$

Performance Metrics (Refined):
Validation MSE:   0.0000616
Validation $R^2$:   0.9881
Test MSE:   0.0003800
Test $R^2$:   0.9491

Figure A12: Analytical functions along with performance metrics for Roughness without RAG generated using the proposed framework for two scenarios (ctx and Eq+ctx).

## Without RAG:

$$\text{Surface Hardness}_{\text{Predicted (Initial)}} = 210.1967 + 4.9672 \cdot B + 0.0665 \cdot R$$
$$- 2.8727 \cdot I + 2.1852 \cdot N^2$$

Performance Metrics (Initial):
Validation MSE:   3.0911
Validation $R^2$:   0.9725
Test MSE:   23.6349
Test $R^2$:   0.9045

$$\text{Surface Hardness}_{\text{Predicted (Refined)}} = 186.8704 + 14.7598 \cdot \ln(B + 1) + 0.4331 \cdot \sqrt{R}$$
$$- 0.2474 \cdot I^2 + 18.3193 \cdot N^{0.5}$$
$$+ 0.0075 \cdot (B \cdot R)$$

Performance Metrics (Refined):
Validation MSE:   2.0315
Validation $R^2$:   0.9819
Test MSE:   9.7301
Test $R^2$:   0.9607

Figure A3: Analytical functions along with performance metrics for Surface Hardness without RAG generated using the proposed framework for two scenarios (ctx and Eq+ctx).

Table A2: Comparison of Performance Metrics (R² score) for Analytical Models Generated With and Without RAG Across Three Manufacturing Processes

### FLIPMM Process

| Process Output | Model | Validation R² score | | Process Output | Model | Validation R² score | |
| --- | --- | --- | --- | --- | --- | --- | --- |
| | | Without RAG | With RAG | | | Without RAG | With RAG |

| Process Output | Model | Without RAG | With RAG | Process Output | Model | Without RAG | With RAG |
|---|---|---|---|---|---|---|---|
| Width | Initial | 0.936 | 0.923 | Depth | Initial | 0.849 | 0.949 |
| | Refined | 0.981 | 0.983 | | Refined | 0.985 | 0.986 |
| Heat Affected Zone (HAZ) | Initial | 0.784 | 0.784 | Material Removal Rate (MRR) | Initial | 0.937 | 0.966 |
| | Refined | 0.982 | 0.980 | | Refined | 0.987 | 0.985 |

SLA Process

| Process Output | Model | Validation R² score | | Process Output | Model | Validation R² score | |
|---|---|---|---|---|---|---|---|
| | | Without RAG | With RAG | | | Without RAG | With RAG |
| Ultimate Tensile Strength (UTS) | Initial | 0.963 | 0.971 | Printing Time | Initial | 0.899 | 0.923 |
| | Refined | 0.981 | 0.983 | | Refined | 0.988 | 0.993 |

TADCR Process

| Process Output | Model | Validation R² score | | Process Output | Model | Validation R² score | |
|---|---|---|---|---|---|---|---|
| | | Without RAG | With RAG | | | Without RAG | With RAG |
| Surface Hardness | Initial | 0.972 | 0.970 | Roughness | Initial | 0.954 | 0.959 |
| | Refined | 0.981 | 0.983 | | Refined | 0.988 | 0.991 |

## A.4 Comparison With PySR

Table A3: Comparison of Performance Metrics (R² score) for Analytical Models from Scenario ctx-refined using proposed framwork and PySR Across Three Manufacturing Processes

| Process Output | Test R² score | | Process Output | Test R² score | |
|---|---|---|---|---|---|
| | PySR | ctx-Refined | | PySR | ctx-Refined |
| Width | 0.844 | 0.958 | Depth | 0.751 | 0.728 |
| Heat Affected Zone (HAZ) | 0.439 | 0.928 | Material Removal Rate (MRR) | 0.668 | 0.901 |
| Process Output | Test R² score | | Process Output | Test R² score | |
| | PySR | ctx-Refined | | PySR | ctx-Refined |
| Ultimate Tensile Strength (UTS) | 0.624 | 0.692 | Printing Time | 0.411 | 0.835 |
| Process Output | Test R² score | | Process Output | Test R² score | |
| | PySR | ctx-Refined | | PySR | ctx-Refined |
| Surface Hardness | 0.933 | 0.948 | Roughness | 0.749 | 0.949 |